\documentclass[11pt,letterpaper]{article}

\usepackage[english]{babel}
\usepackage[T1]{fontenc}
\usepackage[utf8]{inputenc}
\usepackage{lmodern}
\usepackage{microtype}

\usepackage[margin=1in]{geometry}
\usepackage{setspace}

\usepackage{amsmath, amsthm, amssymb, amsfonts}

\usepackage{algorithm}

\usepackage{graphicx}
\graphicspath{{Figure/}}
\usepackage{caption}
\usepackage{float}

\usepackage{booktabs}
\usepackage{multirow}
\usepackage{makecell}

\usepackage{natbib}
\usepackage{xurl}

\usepackage{algpseudocode}

\usepackage{xcolor}
\usepackage{enumerate}

\definecolor{linkblue}{RGB}{24,64,112}
\usepackage[colorlinks=true,allcolors=linkblue,breaklinks=true,pdfencoding=auto,psdextra]{hyperref}
\hypersetup{pdftitle={Censored Flow Matching for Synthetic Time-to-Event Generation},pdfauthor={Jianbin Tan, Xingran Chen, Kevin He, Zhenke Wu, Bhramar Mukherjee, Anru R. Zhang}}
\floatstyle{ruled}
\restylefloat{algorithm}
\theoremstyle{definition}
\newtheorem{Theorem}{\textbf{Theorem}}

\newtheorem{Remark}{Remark}

\newtheorem{Proposition}{Proposition}
\newtheorem{Assumption}{Assumption}

\def\bm{\boldsymbol}
\usepackage{xr}
\title{Censored Flow Matching for Synthetic Time-to-Event Generation}
\author{Jianbin Tan\thanks{Department of Biostatistics \& Bioinformatics, Duke University}, ~ Xingran Chen\thanks{Department of Biostatistics, University of Michigan}, ~Kevin He\thanks{Department of Biostatistics, University of Michigan}, ~ Zhenke Wu\thanks{Department of Biostatistics, University of Michigan},\\ 
~ Bhramar Mukherjee\thanks{Department of Biostatistics, Yale University}, ~and~ Anru R. Zhang\thanks{Department of Biostatistics \& Bioinformatics and Department of Computer Science, Duke University}}
\date{}

\begin{document}

\maketitle

\begin{abstract}
\singlespacing

Generating time-to-event data is increasingly important in survival analysis, but
it remains statistically challenging because event times are often censored and thus prevent flexible generative modeling.
To fill this gap, we propose a novel method called censored flow matching (CFM), aiming at providing a flexible flow-based generative mechanism for synthesizing time-to-event data as well as learning their corresponding probabilistic densities.
Unlike existing methods, CFM does not rely on survival likelihoods to handle censoring, but instead exploits the monotonicity property of differential flows to address censoring via pullback sampling.
Under this framework, we implement an alternating procedure between pullback sampling and CFM steps, leading to a flow-based expectation--maximization (EM) algorithm tailored for time-to-event inference.
Under mild distributional conditions, we establish the convergence of CFM to the true event-time distribution as the EM iterations proceed.
We then apply CFM to generate time-to-event data and estimate their probabilistic densities, where the observed censoring time can be incorporated to provide truncated sampling or truncated density estimation for survival inference.
Through extensive simulation studies, we demonstrate the benefits of CFM for complex and large-scale data generation.
Finally, we apply CFM to analyze data from the Organ Procurement and Transplantation Network (OPTN) and the Medical Information Mart for Intensive Care IV (MIMIC-IV), showcasing its benefits for synthetic and predictive survival analyses in terms of data fidelity, predictive performance, and empirical privacy characteristics.

\end{abstract}
\setstretch{1.08}

\noindent{\it Keywords:} Flow-based expectation--maximization algorithm, flow matching, ordinary differential equation, survival analysis, time-to-event generation

\section{Introduction}

Generating time-to-event data is essential for many real-world applications in survival analysis, including imputing event times or estimating probabilistic densities for individualized prediction, as well as synthesizing event-time datasets when direct access to survival data is restricted. These tasks are fundamental in clinical analysis, where clinical decision often requires individualized inference, uncertainty quantification, and privacy-preserving analysis of time-to-event data. 
However, survival time data are often subject to censoring; that is, for some subjects, the exact event time is not observed, and we only know that the event occurs after the last recorded follow-up time \citep{royston2002flexible,kalbfleisch2002statistical,xu2014standard,apellaniz2024leveraging,su2026nonparametric}.
This phenomenon leads to a missing-data issue in clinical analysis, preventing the direct generation of time-to-event due to the incomplete nature of observed data.
Addressing censoring is therefore a central issue in synthetic survival analysis and has received increasing attention in recent literature \citep{norcliffe2023survivalgan,apellaniz2024leveraging,brockschmidt2025survdiff}.

Broadly speaking, existing event-time generative methods may be summarized into two paradigms.
The first focuses on generating time-to-event data together with censoring indicators, such as \citet{norcliffe2023survivalgan,ashhad2024conditioning,brockschmidt2025survdiff}. This perspective treats the observed data as a tabular dataset \((Y,\Delta,\bm X)\) and generates data in the same form, where \(Y\) is the observed time, \(\Delta\) indicates whether \(Y\) is the true event time \((\Delta=1)\) or a censoring time \((\Delta=0)\), and \(\bm X\) denotes potential covariates.
Accordingly, existing tabular generative methods can be extended to time-to-event generation by incorporating survival mechanisms into training and generation processes \citep{norcliffe2023survivalgan,brockschmidt2025survdiff}.
However, because these approaches generate the survival tuple \((Y,\Delta,\bm X)\) rather than the underlying uncensored event time, the resulting synthetic data reflect uncertainty in both event times and the censoring process. This uncertainty may carry over to and potentially increase variability in downstream analyses.
Moreover, the existing methods are designed to match the observed censored-data distribution rather than directly recover the underlying uncensored distribution, and therefore may not serve the goal of predicting latent event times for censored subjects.

The second paradigm focuses on generating latent uncensored event times \(T\) from censored observations, which is applicable to event-time imputation, individualized prediction, and addressing censoring in synthetic data analysis.
To handle missingness in event time, this paradigm usually combines survival analysis tools with likelihood-based generative methods \citep{xiu2020variational,ausset2021individual}, such as variational autoencoders \citep[VAE;][]{doersch2016tutorial} and neural ordinary differential equations \citep[NODE;][]{chen2018neural}.
However, these methods may have limitations in generative modeling: VAEs typically generate event times through a direct transport from latent variables, where the transport is not guaranteed to be optimal and the  learned latent representations may not match the prespecified latent prior, compromising the fidelity of generated data; NODE, on the other hand, requires adjoint-based computation for likelihood evaluation and model training \citep{chen2018neural}, which can be computationally expensive and less scalable for large datasets.

Recently, state-of-the-art generative methods, including diffusion models \citep{ho2020denoising,yang2023diffusion} and flow matching \citep{liu2022flow,lipman2022flow}, have demonstrated strong empirical performance over VAEs and NODEs \citep{ho2020denoising,lipman2022flow,yang2023diffusion,gat2024discrete,chen2024flow,tan2025smooth}. 
A key advantage of these methods is that they do not directly fit the likelihood, but instead learn a continuous transport path from a base distribution to the data distribution.
By decomposing a potentially complex transport into a sequence of local transformations, diffusion-/flow-based methods make the base-to-target transport easier to learn, enabling expressive representations for distributional learning and data generation.
Furthermore, these methods permit adjoint-free training through a least-squares objective, yielding a computationally efficient and scalable framework for generative modeling of large datasets.

Despite their success, existing continuous generation methods mainly focus on structured tabular data rather than incomplete target generation. This is because training such generative paths usually requires fully observed samples to guide the transport from latent to target samples. For time-to-event data, however, censoring provides only a lower bound on the true event time rather than full guidance for the generative path, making continuous generation difficult to define in this setting.

\begin{figure}
    \centering
    \includegraphics[width=0.99\linewidth]{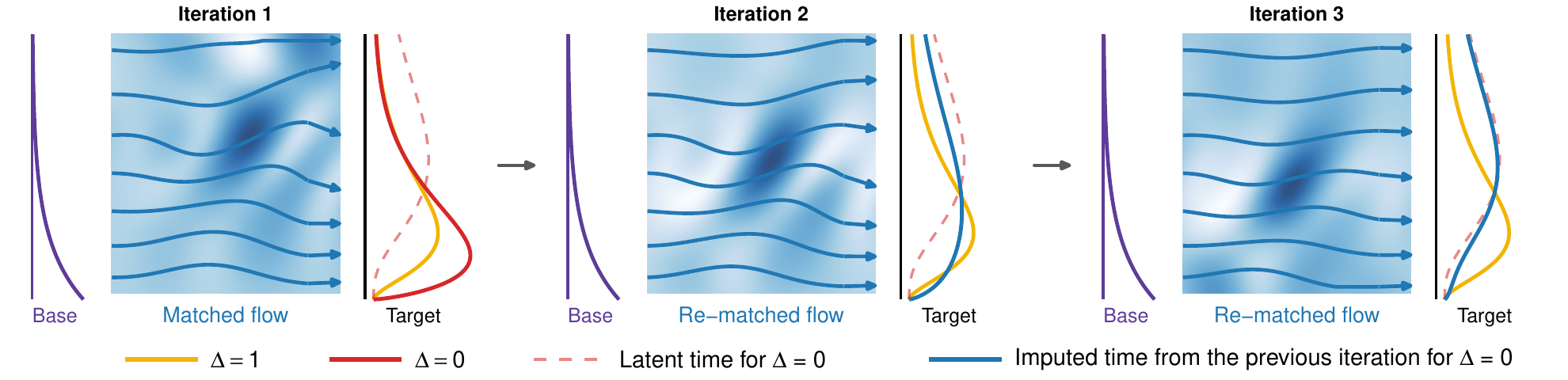}
    \caption{An illustration of the iterative procedure in censored flow matching. The purple, yellow, red, and blue solid curves represent {the density of a base variable (e.g., an exponential random variable)}, the observed time for non-censored subjects \((\Delta=1)\), the observed time for censored subjects \((\Delta=0)\), and the imputed time for censored subjects based on the flow matched in the previous iteration, respectively. The flow at each iteration is obtained by matching the interpolation paths from the base to the target mixture represented by the {blue (or red) and yellow} solid curves.
    The red dashed curve denotes the latent time density for censored subjects, which is approached by the imputed density as the iterations proceed.
    }
    \label{fig:CFM_illu}
\end{figure}

To address these challenges, we develop a novel flow-matching framework for continuously generating uncensored event times from censored survival data.
To handle censoring, we utilize the monotonicity property of ordinary differential equations (ODEs) to construct a {\it censored flow matching} (CFM) step, which enables pullback-flow sampling to impute event times from the observed censoring lower bound.
We then establish an iterative CFM algorithm that alternates between pullback sampling and flow estimation, mimicking the E-step and M-step of an expectation--maximization (EM) algorithm (see Figure~\ref{fig:CFM_illu}).
We prove that this procedure is exactly a flow-based EM algorithm, where both the imputation distribution and the observed-data distribution are modeled through ODE flows.
Under mild distributional conditions, we establish the convergence of CFM across EM iterations, highlighting its nonparametric flexibility for time-to-event generation.

\begin{figure}
    \centering
    \includegraphics[width=\linewidth, trim=0cm 1cm 0cm 0 cm, clip]{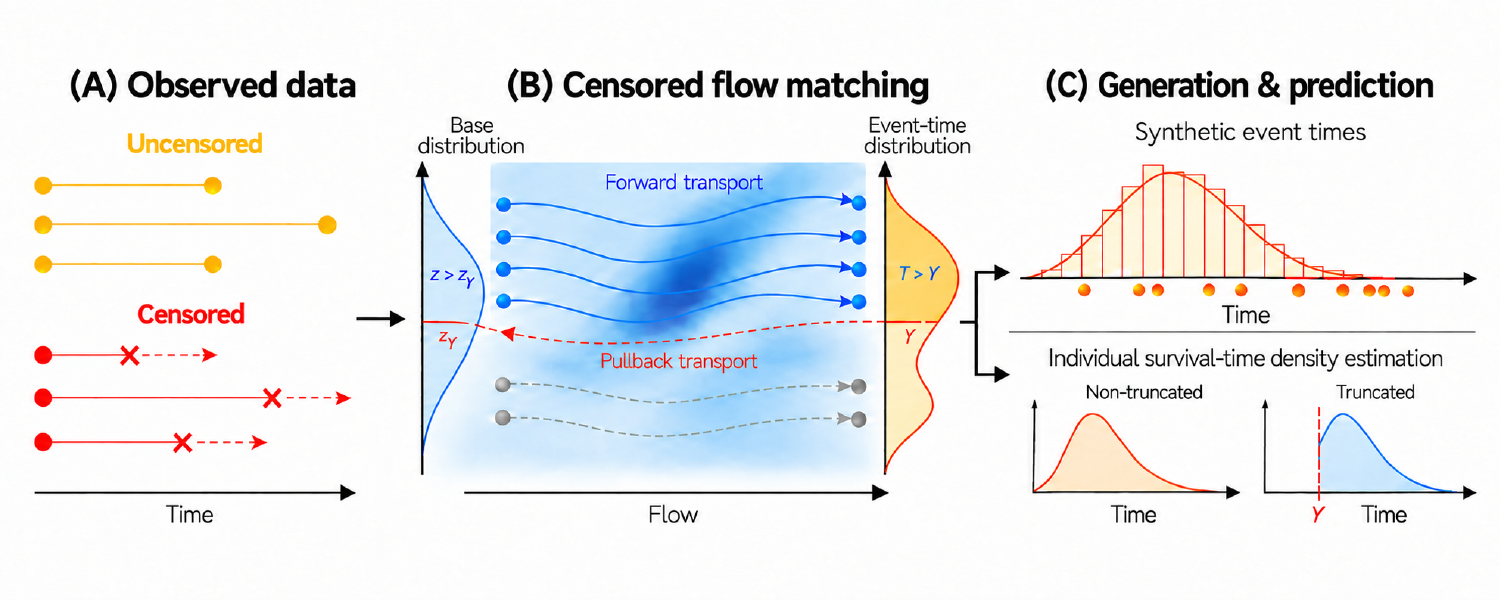}
    \caption{
A pipeline for censored flow matching. Panel (A) illustrates the input data including uncensored and right-censored observations. Panel (B) illustrates the CFM step, where pullback transport guides forward transport to generate event times beyond {the observed time \(Y\) for a censored subject}.
Panel (C) presents the outputs of CFM for synthetic event time generation and individual survival-time density estimation.
}
    \label{fig:CFM_app}
\end{figure}

Building upon this framework, we propose a neural-network-based CFM algorithm for empirical time-to-event data using stochastic minibatch optimization. The algorithm is survival-likelihood-free and adjoint-free during flow training, relying on iterative least-squares fitting guided by the  pullback-flow sampling.
The learned flow provides a continuous mechanism that is guaranteed to be an optimal transport for time-to-event generation and can be used to (i) generate full event times for non-censored subjects or truncated event times for censored subjects given the last recorded follow-up time, and (ii) estimate the corresponding full or truncated densities directly from the flow.
These tasks are fundamental in real-world synthetic and predictive survival analyses.
A pipeline of the proposed CFM is shown in Figure~\ref{fig:CFM_app}.

Through extensive simulations, we show that CFM better generates event times than existing survival generative methods.
In real-data analyses, we apply CFM to the Organ Procurement and Transplantation Network (OPTN) database for synthetic survival data generation and to the Medical Information Mart for Intensive Care IV (MIMIC-IV) database \citep{johnson2020mimic} for survival prediction.
On the OPTN data, CFM more closely reproduces overall and subgroup-specific survival distributions than existing survival generative methods, while exhibiting favorable empirical privacy characteristics.
On the MIMIC-IV data, CFM allows direct distribution-level prediction for censored subjects and achieves competitive prediction accuracy compared with existing survival methods.

\subsection{Related Literature}\label{sec:related lit}

We next review existing survival and generative approaches in connection with CFM.

\paragraph*{Hazard-based Generation.}
In addition to VAE and NODE, hazard-based methods can also be used to generate uncensored time-to-event data by exploiting the relationship between survival time and survival probability.
Specifically, one may first estimate the conditional hazard function
\[
h(t\mid \bm X=\bm x)
=
\lim_{\Delta t\downarrow 0}
\frac{
\mathbb P(t\le T<t+\Delta t\mid T\ge t,\bm X=\bm x)
}{
\Delta t
}
\]
as well as survival function
\(
S(t\mid \bm X=\bm x)
:=
\mathbb{P}(T\geq t\mid \bm{X}=\bm{x})
=
\exp\left\{
-H(t\mid \bm X=\bm x)
\right\},
\)
where
\(
H(t\mid \bm X=\bm x)
=
\int_0^t h(s\mid \bm X=\bm x)\,\mathrm ds
\)
is the cumulative hazard.
The hazard function can be estimated from censored observations using existing survival analysis methods, such as  \citet{cox1975partial,kvamme2019time,tang2022soden,su2026nonparametric}.
Since the cumulative hazard is the negative logarithm of the conditional survival function, event times can be generated by drawing \(U\sim \mathrm{Unif}(0,1)\) and solving
\(
H(\widetilde{T}\mid \bm X=\bm x)=-\log U,
\)
or equivalently,
\begin{eqnarray}\label{remark_hazard}
\widetilde T
=
H^{-1}(-\log U\mid \bm X=\bm x).
\end{eqnarray}
Note that this procedure requires inverting the estimated cumulative hazard function, making generation dependent on hazard estimation accuracy and potentially introducing inversion error. In contrast, CFM directly learns an ODE-based transport from a base random variable (e.g., an exponential random variable) to the event-time data, providing a more direct generative approach.

\paragraph*{Continuous Generation for Incomplete Data.}
Generative modeling for incomplete data commonly arises in clinical and biological applications with irregular, partial, or missing observations.
Existing continuous generative models can accommodate certain forms of incomplete data under suitable conditions.
For example, diffusion models \citep{song2020score,rozet2024learning,zhang2025diffputer} handle ignorable missingness by exploiting that the score of missing components can be derived from the joint score; \citet{zhang2024trajectory} and \citet{tan2026panel} address irregular longitudinal observations in flow matching through structured longitudinal flows; SurvDiff \citep{brockschmidt2025survdiff} tackles censored survival-data generation by incorporating a survival likelihood-based loss into diffusion model training.
However, these approaches do not address the generation of uncensored event times from censored data, where the key challenge is how to learn a continuous generator when the available information only reveals that the event occurs after the observed censoring time.
Here, we develop CFM to address this issue, utilizing the monotonicity of ODE flows to incorporate partial ordering into deterministic generative paths. This property allows censoring times to be translated back into constraints in the base space, enabling an effective forward transport path guided by censored data. The backward-to-forward scheme leads to statistically valid learning of an uncensored event-time generator.

\subsection{Organization}

The rest of this paper is organized as follows. Section~\ref{sec: method} introduces the proposed CFM framework, including the pullback-flow sampling procedure and the EM-type alternating algorithm. Section~\ref{sec: EM} proves that CFM is exactly a flow-based EM algorithm and establishes its convergence properties. Section~\ref{sec: CFMem} presents the implementation of CFM for empirical survival data. Section~\ref{sec:sim} reports simulation studies, and Section~\ref{sec:data} illustrates the proposed method using real data. We conclude with a discussion in Section~\ref{sec:ids}.

\section{Time-to-Event Generation}\label{sec: method}

Throughout this article, \(N(a,b)\) denotes a normal distribution with mean \(a\) and covariance \(b\), and \(\bm I_d\) denotes the \(d\)-dimensional identity matrix. We use ``\(\overset{d}{=}\)'' to denote equality in distribution. We denote the density of a random variable \(X\) by \(p_X\). 
For a density \(p\) and a random variable \(X\), we write
\(
\operatorname{supp}(p)
\)
and
\(
\operatorname{supp}(X)
\),
respectively, for their supports.
Let \( \mathbb{I}(\cdot) \) denote the indicator function and \(\|\cdot\|\) denote the Euclidean norm. 

Let \(\bm X\in\mathcal{X}\) denote a vector of covariates, and let \(T\mid \bm X \sim p_T(\cdot\mid \bm X)\) denote the event time of interest, where $\mathcal{X}$ is a subset of $\mathbb R^d$.
Let \(C\in\mathbb R_{+}\) denote another random time. We assume that \(C\) and \(T\) are conditionally independent given \(\bm X\).
Under right censoring, we observe
$Y=\min(T,C),\ 
\Delta=\mathbb{I}(T\le C),$
where \(Y\) is the observed time and \(\Delta\) is the censoring indicator, with \(\Delta=1\) indicating that the event time is observed and \(\Delta=0\) indicating right censoring. 

We first develop CFM at the population level, assuming that the joint distribution of $(Y,\Delta,\bm X)$ is known and can be sampled from. Section~\ref{sec: CFMem} then considers the empirical setting based on an i.i.d. sample $\{(Y_i,\Delta_i,\bm X_i):i=1,\ldots,n\}$.
At the population level, our target is a continuous flow that, conditional on $\bm X$, generates an event time $\widetilde T$ satisfying
$\widetilde T\mid \bm X \overset{d}{=} T\mid \bm X$.

\subsection{Flow Matching for Censored Data}

In this subsection, we introduce a flow-matching framework for generating event-time from \((Y,\Delta,\bm X)\).
Before presenting our approach, we review standard flow matching given non-censored event-time data \((T,\bm X)\). 
Let \(Z\sim p_Z\) denote a base random variable with known density \(p_Z\) that is easy to sample from.
The goal of flow matching is to learn a covariate-dependent vector field \(v\) that transports samples from \(p_Z\) to the target conditional distribution \(p_T(\cdot\mid \bm X)\).
Specifically, after learning \(v\), a new event-time conditional on a covariate value \(\bm X=\bm x\) is generated by first sampling \(Z_0\sim p_Z\), and then solving the ODE
\begin{eqnarray}\label{ode_generation}
    \frac{\mathrm d Z_u}{\mathrm d u}
=
v(u,Z_u,\bm x),
\qquad u\in[0,1],
\end{eqnarray}
from \(u=0\) to \(u=1\); the generated event time is then given by
\(
\widetilde T=Z_1.
\)
This generation mechanism suggests that the distribution of $\widetilde T$ conditioning on $\bm{X}$ is induced by $p_Z$ and the vector field $v$.

An efficient way to estimate the vector field \(v\) is through rectified flow \citep{liu2022flow}, which learns the velocity along straight-line interpolations between samples from \(p_Z\) and the target distribution. Specifically, for \(u\in[0,1]\), define \(T_u=(1-u)Z+uT\). Since \(T_0\sim p_Z\) and \(T_1\sim p_T(\cdot\mid\bm X)\), the path \(\{T_u:u\in[0,1]\}\) interpolates from the base to the target conditional distribution. Moreover, \(\frac{\mathrm dT_u}{\mathrm du}=T-Z\), so rectified flow matching seeks a vector field \(v\) such that \(v(u,T_u,\bm X)\approx T-Z\), consistent with the ODE generation mechanism in \eqref{ode_generation}. The population flow-matching loss over a vector field $v$ is then given by
\begin{eqnarray}\label{loss_FM}
\mathcal L_{\mathrm{FM}}(v)
=
\mathbb E_{T,Z,U,\bm X}
\left[
\left\{
v(U,T_U,\bm X)-(T-Z)
\right\}^2
\right],
\end{eqnarray}
where \(T\sim p_T(\cdot\mid\bm X)\), \(Z\sim p_Z\), \(U\sim\mathrm{Unif}(0,1)\), and \(\bm X\sim p_{\bm X}\). In practice, the expectation is approximated using empirical samples and $v$ is parameterized by a neural network. Minimizing \(\mathcal L_{\mathrm{FM}}(v)\) over $v$ then learns a vector field that transports samples from \(p_Z\) to \(p_T(\cdot\mid\bm X)\) \citep{liu2022flow,lipman2022flow,lipman2024flow}.

\paragraph*{Censored Flow Matching.}
We now extend flow matching to the censored data \((Y,\Delta,\bm X)\). Let \(q\) be an arbitrary conditional density of \(T\) among censored observations.
The CFM loss is defined as
\begin{eqnarray*}
\mathcal L_{\mathrm{CFM}}(v,q)
&=&
\mathbb{P}(\Delta = 1)\,
\mathbb E
\left[
\left.
\left\{
v(U,Y_U,\bm X)-(Y-Z)
\right\}^2
\,\right|\, \Delta = 1
\right]
\\
&&+
\mathbb{P}(\Delta = 0)\,
\mathbb E\bigg\{ 
\mathbb E_{T'\sim q(\cdot\mid \Delta=0,Y,\bm X)}
\left[
\left.
\left\{
v(U,T'_U,\bm X)-(T'-Z)
\right\}^2
\,\right|\, \Delta = 0, Y, \bm X
\right]\bigg\},
\end{eqnarray*}
where
\(
Y_u=(1-u)Z+uY\) and \(T'_u=(1-u)Z+uT'\), $u \in [0,1]$.
Intuitively, for uncensored observations with \(\Delta=1\), the event time is observed, and we have \(Y=T\), as in the standard flow-matching loss.
For censored observations with \(\Delta=0\), the true event time is unobserved but satisfies \(T>Y\); therefore, we replace it with a completed event time \(T'\) sampled from \(q(\cdot\mid \Delta=0,Y,\bm X)\), where \(q\) is an imputation distribution for the latent event time.

The following result establishes the equivalence between CFM and FM under true laws.

\begin{Proposition}[Equivalence between CFM and FM]\label{pro_1}
Suppose that the imputation distribution \(q\) equals the true conditional distribution of the latent event time, that is,
\(
    q(\cdot\mid \Delta=0,Y,\bm X)
    =
    p_T(\cdot\mid \Delta=0,Y,\bm X).
\)
Then, for any vector field \(v\) with
\(\mathcal L_{\mathrm{FM}}(v)<\infty\),
\(
    \mathcal L_{\mathrm{CFM}}(v,q)
    =
    \mathcal L_{\mathrm{FM}}(v).
\)
\end{Proposition}

Proposition~\ref{pro_1} shows that the CFM loss equals the complete-data FM loss when \(q\) is the true conditional distribution of the latent event time. This parallels the EM principle, where the observed-data likelihood is obtained by marginalizing the complete-data likelihood over missing components conditional on the observed data.
This motivates an EM-type iterative flow-matching procedure, implemented by alternating between updating \(q\) and minimizing the CFM loss.

Specifically, given a current working conditional density \(q^{(k)}(\cdot\mid \Delta=0,Y,\bm X)\) at the \(k\)th step, we update the vector field by solving
\begin{eqnarray}\label{iter_up}
    v^{(k+1)}
=
\arg\min_{v}
\mathcal L_{\mathrm{CFM}}(v,q^{(k)}).
\end{eqnarray}
Let \(\{(Y_b,\Delta_b,\bm X_b):b=1,\ldots,B\}\) be a batch of
\(B\) independent copies of \((Y,\Delta,\bm X)\).
The update \eqref{iter_up} can then be implemented by approximating the population loss with the empirical least-squares loss
\[
    \widehat{\mathcal L}_{\mathrm{CFM}}(v,q^{(k)})
    =
    \frac{1}{B}
    \sum_{b=1}^B
    \left[
    v\left(U_b,\widetilde T_{b,U_b}^{(k)},\bm X_b\right)
    -
    \left(\widetilde T_b^{(k)}-Z_b\right)
    \right]^2,
\]
where \(U_b\sim \mathrm{Unif}(0,1)\), \(Z_b\sim p_Z\), \(
    \widetilde T_{b,U_b}^{(k)}
    =
    (1-U_b)Z_b+U_b\widetilde T_b^{(k)}
\), and \(
    \widetilde T_b^{(k)}
    =
    \Delta_bY_b+(1-\Delta_b)T_b^{(k)}\), with  
    \(T_b^{(k)}
    \sim
    q^{(k)}(\cdot\mid \Delta_b=0,Y_b,\bm X_b).
\)
Then, using the updated vector field \(v^{(k+1)}\), we refine the imputation density for the censored observations to obtain
\(
q^{(k+1)}(\cdot\mid \Delta=0,Y,\bm X),
\)
which is used in the next CFM loss evaluation.
The procedure produces a sequence of updated vector fields, and the final vector field is used to generate new synthetic event times. The complete procedure is summarized in Algorithm~\ref{alg:cfm}. 

\begin{algorithm}[htbp]
\caption{Censored Flow Matching (CFM)}
\label{alg:cfm}
\begin{algorithmic}[1]
\Require Data generator for \((Y,\Delta,\bm X)\), base distribution \(p_Z\), initial conditional density \(q^{(0)}(\cdot\mid \Delta=0,Y,\bm X)\), number of iterations \(K\), batch size \(B\).
\Ensure
A new event-time sample \({\widetilde T}^{\mathrm{new}}\) conditioned on \(\bm X^{\mathrm{new}}\).

\For{\(k=0,\ldots,K-1\)}
    \State Generate \(B\) independent samples
    \(
        \{(Y_b,\Delta_b,\bm X_b):b=1,\ldots,B\}
    \)
    from the data generator.

    \For{\(b=1,\ldots,B\)}
       \State Sample \(Z_b\sim p_Z\), \(U_b\sim \mathrm{Unif}(0,1)\), and
        \(
            T_b^{(k)}
            \sim
            q^{(k)}(\cdot\mid \Delta_b=0,Y_b,\bm X_b)
        \) if $\Delta_b=0$.
        Set
        \[
            \widetilde T_b^{(k)}
            =
            \Delta_bY_b+(1-\Delta_b)T_b^{(k)},
            \qquad
            \widetilde T_{b,U_b}^{(k)}
            =
            (1-U_b)Z_b+U_b\widetilde T_b^{(k)}.
        \]
    \EndFor

    \State Update the vector field by minimizing
    \[
        v^{(k+1)}
        =
        \arg\min_{v}
        \frac{1}{B}
        \sum_{b=1}^B
        \left[
        v\left(U_b,\widetilde T_{b,U_b}^{(k)},\bm X_b\right)
        -
        \left(\widetilde T_b^{(k)}-Z_b\right)
        \right]^2.
    \]

    \State Update
    \(
        q^{(k+1)}(\cdot\mid \Delta=0,Y,\bm X)
    \) based on $v^{(k+1)}$.
\EndFor

\State Generate a new event time by sampling \(Z_0\sim p_Z\) and solving ODE~\eqref{ode_generation} with \(v=v^{(K)}\) and \(\bm{x}=\bm X^{\mathrm{new}}\), yielding \({\widetilde{T}}^{\mathrm{new}}\).

\end{algorithmic}
\end{algorithm}

\subsection{Pullback Flow for Censored Sampling}

To implement Algorithm~\ref{alg:cfm}, we need to sample \(T'\) from the density
\(
q^{(k)}(\cdot\mid \Delta=0,Y,\bm X=\bm{x})
\)
which is linked with \(v^{(k)}\).
In particular, for a censored observation with \(\Delta=0\), the latent event time must satisfy \(T>Y\). This means that \(q^{(k)}\) is obtained by restricting the event-time distribution induced by \(v^{(k)}\) to the region \((Y,\infty)\). 
Our goal is therefore to generate \(T'\) as \(Z_1\) subject to the constraint \(Z_1>Y\), where \(Z_1\) is the output of the ODE \eqref{ode_generation} with \(v=v^{(k)}\), \(Z_0\sim p_Z\), and $\bm{x}$. In the following, we show that this can be accomplished by a pullback-flow approach. 

\begin{Theorem}[Pullback-Flow Sampling]\label{Theo:for_back}
Assume that the vector field
\(v^{(k)}\) at the \(k\)th step generates a unique global flow; that
is, for every \(w\in[0,1]\) and \(z\in\mathbb R\), the initial-value
problem
\(
    \frac{\mathrm dZ_u}{\mathrm du}
    =
    v^{(k)}(u,Z_u,\bm x),
    \
    Z_w=z,
\)
admits a unique solution on \(u\in [0,1]\). For a given \(\bm X=\bm x\) and censored time \(Y\), define \(z_Y=Z^{\mathrm{inv}}_1\), where $\{Z^{\mathrm{inv}}_u;u\in [0,1]\}$ is the solution of the backward ODE
\begin{eqnarray}\label{back_flow}
    \frac{\mathrm dZ^{\mathrm{inv}}_u}{\mathrm du}
    =
    -v^{(k)}(1-u,Z^{\mathrm{inv}}_u,\bm x),
    \qquad u\in[0,1],
    \qquad Z^{\mathrm{inv}}_0=Y.
\end{eqnarray}
Then the constraint \(Z_1>Y\) under the forward
ODE~\eqref{ode_generation} is equivalent to the constraint
\(Z_0>z_Y\) in the base space, i.e.,
\begin{eqnarray}\label{pro_equi}
    \mathbb P\!\left(
        Z_1>y
        \mid
        \bm X=\bm x
    \right)
    =
    \mathbb P\!\left(
        Z_0>z_y
    \right),
    \qquad y\in\mathbb R.
\end{eqnarray}
Moreover, 
a sample from the conditional distribution of \(Z_1\) given
\(Z_1>Y\) and \(\bm X=\bm x\) can be obtained by first sampling
\(
    Z_0\sim p_Z(\cdot\mid Z_0>z_Y)
\)
and then solving the forward ODE~\eqref{ode_generation} with
\(v=v^{(k)}\) from \(u=0\) to \(u=1\).
\end{Theorem}

Theorem~\ref{Theo:for_back} shows that sampling \(Z_1>Y\) using \(v^{(k)}\) can be achieved by first computing the pullback value \(z_Y\), then sampling the base variable from the truncated distribution \(p_Z(\cdot\mid Z_0>z_Y)\), and finally propagating the sampled \(Z_0\) forward through the ODE. This property follows from the monotonicity of the one-dimensional flow; that is, the flow map induced by \eqref{ode_generation}, denoted by \(\Phi_1^{(k)}(\cdot;\bm{x})\), is strictly increasing. Here, the flow \(\{\Phi^{(k)}_u(\cdot;\bm{x}):u\in[0,1]\}\) associated with \(v^{(k)}\) solves
\[
\frac{\mathrm{d}\Phi_u^{(k)}(z;\bm x)}{\mathrm{d}u}
=
v^{(k)}(u,\Phi_u^{(k)}(z;\bm x),\bm x),
\qquad
\Phi_0(z;\bm x)=z.
\]
See Part~\ref{proof_mono} in the Supplementary Materials for a proof. Under this property, the truncation region in the target space can be pulled back to the corresponding region in the base space, making sampling from the desired truncated distribution straightforward.

\begin{Remark}[Truncated Density Evaluation]
Let \(p_{T,v^{(k)}}\) be the
density generated by \(p_Z\) and \(v^{(k)}\) via ODE~\eqref{ode_generation}.
Theorem~\ref{Theo:for_back} can be used to obtain truncated conditional
densities for \(p_{T,v^{(k)}}\) given a truncation point \(Y=y\ge 0\). 
For \(t\in(y,\infty)\), define
\begin{eqnarray}\label{con_equ}
     p_{T,v^{(k)}}(t\mid \bm X=\bm x,T>y)
    :=
    \frac{
        p_{T,v^{(k)}}(t\mid \bm x)\mathbb I(y<t)
    }{
        \int_y^\infty p_{T,v^{(k)}}(r\mid \bm x)\,\mathrm{d}r
    } .
\end{eqnarray}
The above truncated conditional density can be evaluated through the
change-of-variables formula along the inverse flow. 
Specifically, for a target value \(t>y\), let \(\{Z_u^{\mathrm{inv}};u\in [0,1]\}\) solve the
inverse ODE \eqref{back_flow}, initialized at
\(Z_0^{\mathrm{inv}}=t\).  Denote \(z_y\) as the pullback value of the truncation
point \(y\), as defined in Theorem~\ref{Theo:for_back}. Under this, we
have
\begin{eqnarray}
\label{densi}
\log p_{T,v^{(k)}}(t\mid \bm X=\bm x,T>y)
&=&
\log p_Z(Z_1^{\mathrm{inv}})
-
\log \mathbb P(Z_0>z_y)
-
\int_0^1
\partial_z v^{(k)}(u,Z_{1-u}^{\mathrm{inv}},\bm x)\,\mathrm{d}u.
\end{eqnarray}
See
Remark~\ref{re_den} in the Supplementary Materials for the detailed proof.
If \(y=0\), \eqref{densi} reduces to the non-truncated conditional density.
\end{Remark}

\begin{Remark}[Optimal Transport Interpretation]
The flow map in \eqref{ode_generation} also admits an optimal transport interpretation, yielding a parsimonious representation of the conditional event-time distribution for survival modeling.
Under the conditions of Theorem~\ref{Theo:for_back},
$\Phi_1^{(k)}(\cdot;\bm x)$ is monotone and transports the base distribution
$p_Z$ to the event-time distribution $p_{T,v^{(k)}}(\cdot\mid\bm X=\bm x)$.
Consequently, for any strictly convex and continuous cost
$c:\mathbb R\to\mathbb R_+$, it solves
\[
    \inf_{G\in\mathcal G_{\bm x}(p_{T,v^{(k)}})}
    \int c\{z-G(z;\bm x)\}\,p_Z(z)\,\mathrm dz,
\]
where $\mathcal G_{\bm x}(p_{T,v^{(k)}})$ denotes the collection of measurable
maps transporting $p_Z$ to
$p_{T,v^{(k)}}(\cdot\mid\bm X=\bm x)$. Thus, the generative flow
$\Phi_1^{(k)}$ coincides with the one-dimensional monotone optimal transport
map. See Remark~\ref{rem_opt} in the Supplementary Materials for the proof.
\end{Remark}

In the following, we specify conditions under which the conditions in
Theorem~\ref{Theo:for_back} holds throughout the iterations. 

\begin{Assumption}\label{as:completed_interp}
Assume that the following conditions hold for fixed \(\bm x\) and
some constant \(I>0\).
\begin{itemize}
    \item[(a)] The base density \(p_Z\) is continuous on \([0,I]\),
    has support \([0,I]\), and is strictly positive on \((0,I)\).

    \item[(b)] The conditional distribution of
    \(Y\mid\Delta=0,\bm X=\bm x\) has a density with support contained
    in \([0,I]\). Moreover,
    \(p_Y(\cdot\mid\Delta=1,\bm X=\bm x)\) is continuous on
    \([0,I]\), has support \([0,I]\), and is strictly positive on
    \((0,I)\).

    \item[(c)] The uncensoring probability satisfies
    \(
        0
        <
        \pi_1(\bm x)
        :=
        \mathbb P(\Delta=1\mid\bm X=\bm x)
        <
        1.
    \)
\end{itemize}
\end{Assumption}

Assumption~\ref{as:completed_interp} imposes mild support,
continuity, and positivity conditions on the base distribution and
the observed-time distributions. Under these conditions, we have the
following result.

\begin{Theorem}[Iterative Regularity Preservation]
\label{the:regularity_v_update}
Suppose Assumption~\ref{as:completed_interp} holds, and fix
\(\bm x\). Assume that \(v^{(k)}\) satisfies the conditions in
Theorem~\ref{Theo:for_back}, with the state space of $z$ restricted to
\([0,I]\). Assume that the terminal distribution induced by \(p_Z\)
and \(v^{(k)}\) has a continuous density
\(p_{T,v^{(k)}}(\cdot\mid\bm x)\) with support \([0,I]\), satisfying
\begin{eqnarray}\label{condition_v}
    \inf_{y\in
    \operatorname{supp}(Y\mid\Delta=0,\bm X=\bm x)}
    \int_y^I
    p_{T,v^{(k)}}(r\mid\bm x)\,\mathrm dr
    >
    0.
\end{eqnarray}
Let \(v^{(k+1)}\) be a population minimizer of \eqref{iter_up} with
\(
    q^{(k)}(\cdot\mid\Delta=0,Y,\bm X)
    =
    p_{T,v^{(k)}}(\cdot\mid\bm X,T>Y).
\)
Then \(v^{(k+1)}\) generates a unique global
flow on \([0,I]\), and hence satisfies the conditions in
Theorem~\ref{Theo:for_back} with the state space of $z$ restricted to
\([0,I]\).
Moreover, the terminal density
\(p_{T,v^{(k+1)}}(\cdot\mid\bm x)\) is continuous on
\([0,I]\), has support \([0,I]\), and satisfies
\(
    \inf_{y\in
    \operatorname{supp}(Y\mid\Delta=0,\bm X=\bm x)}
    \int_y^I
    p_{T,v^{(k+1)}}(r\mid\bm x)\,\mathrm dr
    >
    0.
\)
\end{Theorem}

Here, the condition
\(
    \inf_{y\in
    \operatorname{supp}(Y\mid\Delta=0,\bm X=\bm x)}
    \int_y^I
    p_{T,v^{(k)}}(r\mid\bm x)\,\mathrm dr
    >
    0
\)
guarantees that \eqref{con_equ} is well defined and prevents the
pullback-flow sampling procedure from relying on vanishing tail
probabilities.

Theorem~\ref{the:regularity_v_update} shows that the proposed iterative CFM
update is stable with respect to the regularity conditions.
In particular, if
\(v^{(k)}\) generates a unique global flow and its terminal density
is continuous, has support \([0,I]\), and satisfies the tail
condition above, then \(v^{(k+1)}\) also admits a unique global flow
and its terminal density satisfies the same conditions. Hence, the
pullback-flow completion step remains well defined and can be applied
recursively.

\section{CFM as a Flow-based EM Algorithm}\label{sec: EM}

In this section, we show that Algorithm~\ref{alg:cfm} is a flow-based EM procedure for time-to-event generation, and establish that, as the iterations proceed, the distribution of the generated event times converges to the true conditional event-time density.

\subsection{\texorpdfstring{\(Q\)}{Q}-Function Maximization via CFM}
In a standard EM algorithm, each iteration consists of an E-step and an M-step.
Typically, the E-step constructs a \(Q\)-function by taking the conditional
expectation of the complete-data log-likelihood with respect to the current
imputation distribution. For the censored data \((Y,\Delta,\bm X)\) and a given
current vector field \(v^{(k)}\), we define the population-level \(Q\)-function as
\begin{eqnarray*}
Q(p\mid p_{T,v^{(k)}})
&=&
\mathbb E
\bigg[
\Delta \log p(Y\mid \bm X)
+
(1-\Delta)
\int_Y^\infty
\log p(t\mid \bm X)\cdot 
p_{T,v^{(k)}}(t\mid \bm X,T>Y)
\,\mathrm dt
\bigg].
\end{eqnarray*}
Under this formulation, the M-step updates the density by maximizing this
\(Q\)-function, i.e.,
\begin{eqnarray*}
     p_{\text{EM}}^{(k+1)}
=
\arg\max_{p}
Q(p\mid p_{T,v^{(k)}}).
\end{eqnarray*}
This constructs an EM algorithm with the imputation distribution modeled by a flow. 

The following result shows that the M-step is equivalent to obtaining
\(v^{(k+1)}\) according to \eqref{iter_up}; that is, CFM is a flow-based EM algorithm in which both the imputation distribution and the observed-data distribution are modeled through ODE flows.

\begin{Theorem}[CFM as an M-step Update]
\label{Theo:completed_target}
Suppose that, for almost every \(\bm x\in\mathcal X\), the
assumptions of Theorem~\ref{the:regularity_v_update} hold at the
\(k\)th iteration. Let \(v^{(k+1)}\) be a population minimizer of
\(\mathcal L_{\mathrm{CFM}}(v,q^{(k)})\) over \(v\), with
\(
    q^{(k)}(\cdot\mid\Delta=0,Y,\bm X)
    =
    p_{T,v^{(k)}}(\cdot\mid\bm X,T>Y).
\)
Then, for almost every
\(\bm x\in\mathcal X\),
\[
    p_{\mathrm{EM}}^{(k+1)}(\cdot\mid\bm X=\bm x)
    =
    p_{T,v^{(k+1)}}(\cdot\mid\bm X=\bm x).
\]
Define the population-level observed log-likelihood for a conditional
density \(p\) by
\(
    \ell_{\mathrm{obs}}(p)
    =
    \mathbb E
    \left[
        \Delta\log p(Y\mid\bm X)
        +
        (1-\Delta)\log S_p(Y\mid\bm X)
    \right],
\)
where
\(
    S_p(Y\mid\bm X)
    =
    \int_Y^\infty p(t\mid\bm X)\,\mathrm dt.
\)
We have
\[
    \ell_{\mathrm{obs}}
    \left(p_{T,v^{(k+1)}}\right)
    \ge
    \ell_{\mathrm{obs}}
    \left(p_{T,v^{(k)}}\right).
\]
\end{Theorem}

Theorem~\ref{Theo:completed_target} shows that the flow-based EM algorithm can be implemented through iterative CFM updates. Consequently, CFM follows the same convergence pathway as the EM algorithm, and we can iteratively update \(v^{(k)}\) in \eqref{iter_up} to increase the observed likelihood.

The above result has important implications. First, CFM provides survival-likelihood-free inference for event-time density estimation, modeling the density through a continuous flow with nonparametric flexibility. This differs from existing likelihood-based survival methods, which often impose parametric models or require explicit likelihood evaluation \citep{guo1992estimating,royston2002flexible,kalbfleisch2002statistical,xu2014standard,apellaniz2024leveraging,su2026nonparametric}. Second, because CFM iteratively updates the flow using least squares, it avoids computationally intensive adjoint methods \citep{chen2018neural} required in existing flow-based survival methods \citep{ausset2021individual,tang2022soden}, making it more scalable to large datasets.

\subsection{Convergence of CFM}
We next establish the convergence of CFM as the iteration proceeds.
For constants
\(c,C_t,C_x,\delta>0\), define the density class
\[
\mathcal M
=
\left\{
p:[0,I]\times\mathcal X\rightarrow\mathbb R:
\begin{aligned}
&\int_0^I p(t\mid\bm x)\,\mathrm dt=1,\quad
p(t\mid\bm x)\ge c,\\
&|p(t\mid\bm x)-p(s\mid\bm x')|
\le
C_t|t-s|+C_x\|\bm x-\bm x'\|,\\
&\inf_{y\in\operatorname{supp}(Y\mid\Delta=0,\bm X=\bm x)}
\int_y^I p(r\mid\bm x)\,\mathrm dr
\ge\delta.
\end{aligned}
\right\}.
\]
Note that $\mathcal{M}$ is a convex set of candidate densities; that is, for any
$p_1,p_2\in\mathcal{M}$ and $\lambda\in[0,1]$,
$\lambda p_1 + (1-\lambda)p_2 \in \mathcal{M}$.
Here, the
uniform lower bound prevents the logarithmic terms in the observed
likelihood and the \(Q\)-function from becoming singular; the joint
Lipschitz condition provides uniform regularity in both the event time
and the covariates;
the uniform tail bound ensures that the truncated conditional
density is well-defined, motivated by the regularity conditions in
\eqref{condition_v}.

For a function \(h\) on \([0,I]\times\mathcal X\), let
\(\|h\|_\infty:=\sup_{t\in[0,I],\,\bm x\in\mathcal X}|h(t\mid\bm x)|\).
We next propose conditions under which \(p_{T,v^{(k)}}\) lies in some
\(\mathcal M\) throughout the updates, as needed for CFM convergence.

\begin{Assumption}[Stability of the CFM update]
\label{as:uniform_cfm_stability}
Suppose that \(\mathcal X\) is compact and that \(p_T\) and
\(p_{T,v^{(0)}}\) are conditional densities supported on \([0,I]\).
Assume that the following conditions hold.
\begin{itemize}

\item[(a)] Define
\(g_1(t\mid\bm x):=\pi_1(\bm x)p_Y(t\mid\Delta=1,\bm X=\bm x)\)
and
\(g_0(t\mid\bm x):=\{1-\pi_1(\bm x)\}p_Y(t\mid\Delta=0,\bm X=\bm x)\).
The functions \(p_T\), \(p_{T,v^{(0)}}\), \(g_0\), and \(g_1\) are
uniformly Lipschitz continuous on \([0,I]\times\mathcal X\).

\item[(b)] Suppose that
\(\inf_{t\in[0,I],\,\bm x\in\mathcal X}p_{T,v^{(0)}}(t\mid\bm x)>0\)
and
\(\inf_{t\in[0,I],\,\bm x\in\mathcal X}g_1(t\mid\bm x)>0\).
Moreover,
\(\inf_{\bm x\in\mathcal X,\,y\in\operatorname{supp}(Y\mid\Delta=0,\bm X=\bm x)}
\int_y^I p_{T,v^{(0)}}(r\mid\bm x)\,\mathrm dr>0\)
and
\(\inf_{\bm x\in\mathcal X,\,y\in\operatorname{supp}(Y\mid\Delta=0,\bm X=\bm x)}
\int_y^I g_1(r\mid\bm x)\,\mathrm dr>0\).

\end{itemize}
\end{Assumption}

Assumption~\ref{as:uniform_cfm_stability} imposes regularity and nondegeneracy conditions on the initial density and observed data. Condition~(a) requires uniform Lipschitz continuity of the event-time density, initial density, and observed-data subdensities. Condition~(b) ensures that the initial density and truncation denominators are uniformly bounded away from zero. 

Accordingly, we define the concept of a stationary point of
\(\ell_{\mathrm{obs}}\) over \(\mathcal M\). We say that
\(p^\star\in\mathcal M\) is a first-order stationary point of
\(\ell_{\mathrm{obs}}\) if the one-sided directional derivative of
\(\ell_{\mathrm{obs}}\) at \(p^\star\) along every feasible direction is
non-positive. That is, for every \(p\in\mathcal M\),
\[
D\ell_{\mathrm{obs}}(p^\star)[p-p^\star]
:=
\left.
\frac{\mathrm d}{\mathrm d\varepsilon}
\ell_{\mathrm{obs}}\bigl(p^\star+\varepsilon(p-p^\star)\bigr)
\right|_{\varepsilon=0+}
\le 0.
\]
Note that the maximizer of \(\ell_{\mathrm{obs}}\) over \(\mathcal M\) is
always a stationary point.

Let \(\widetilde T_k\) be the event time conditional on
\(\bm X\), which is generated by \(v^{(k)}\) and \(p_Z\) through
ODE~\eqref{ode_generation}. Define the conditional
Wasserstein distance between \(\widetilde T_k\) and \(T\) given
\(\bm X=\bm x\) by
\[
    W_2\!\left(\widetilde T_k,T\mid \bm X=\bm x\right)
    =
    \left\{
    \inf_{\gamma\in\Gamma(\bm x)}
    \int_{[0,I]\times[0,I]}
    |t-s|^2
    \,\mathrm d\gamma(t,s)
    \right\}^{1/2},
\]
where \(\Gamma(\bm x)\) is the set of all couplings of
\(\widetilde T_k\mid \bm X=\bm x\) and \(T\mid \bm X=\bm x\). 
We examine the convergence of CFM for data generation and truncated density estimation as follows. 

\begin{Theorem}[Convergence of CFM]
\label{the:obs_likelihood_convergence}
Suppose Assumptions~\ref{as:completed_interp} and
\ref{as:uniform_cfm_stability} hold, and suppose that
\(
\ell_{\mathrm{obs}}(p_{T,v^{(0)}})>-\infty.
\)
Then the following conclusions hold:
\begin{itemize}

\item[(a)] There exist constants \(c,\delta>0\) and
\(C_t,C_x<\infty\) such that \(p_T\) and
\(
\{p_{T,v^{(k)}}\}_{k\ge0}
\)
are contained in the corresponding density class \(\mathcal M\).

\item[(b)] The true density \(p_T\) is a global maximizer of
\(\ell_{\mathrm{obs}}\) over \(\mathcal M\), and every limit point of
\(
\{p_{T,v^{(k)}}\}_{k\ge0}
\)
with respect to \(\|\cdot\|_\infty\) is a stationary point of
\(\ell_{\mathrm{obs}}\) over \(\mathcal M\).

\item[(c)] If \(p_T\) is the unique stationary point of
\(\ell_{\mathrm{obs}}\) over \(\mathcal M\), then, as \(k\to\infty\),
\[
\begin{gathered}
\ell_{\mathrm{obs}}\left(p_{T,v^{(k)}}\right)
\longrightarrow
\ell_{\mathrm{obs}}(p_T),\\
W_2\left(\widetilde T_k,T\mid\bm X=\bm x\right)
\longrightarrow 0,\\
\sup_{y\in\mathcal Y_{\bm x}}
\sup_{t\in(y,I]}
\left|
\log p_{T,v^{(k)}}(t\mid\bm X=\bm x,T>y)
-
\log p_T(t\mid\bm X=\bm x,T>y)
\right|
\longrightarrow 0,
\end{gathered}
\]
where \(\mathcal Y_{\bm x}\) is any set satisfying
\(
\inf_{y\in\mathcal Y_{\bm x}}
\int_y^I p_T(s\mid\bm x)\,\mathrm ds>0.
\)

\end{itemize}
\end{Theorem}

Theorem~\ref{the:obs_likelihood_convergence}(b) shows that \(p_T\) is a
stationary point of the likelihood, and the CFM sequence
\(
    \left\{
        p_{T,v^{(k)}}
    \right\}_{k\ge0}
\)
may potentially approach this stationary point as the iterations proceed.
Here, (b) is established by
equipping \(\mathcal M\) with the topology induced by the norm
\(
    \|\cdot\|_\infty.
\)
Under this topology, \(\mathcal M\) is compact, and both
\(\ell_{\mathrm{obs}}(p)\) and \(Q(p\mid q)\) are continuous with respect to
densities \(p,q\) in \(\mathcal M\).
These properties are key to establishing the convergence of an EM-type algorithm
\citep{wu1983convergence}.

Theorem~\ref{the:obs_likelihood_convergence}(c) shows that CFM can recover the maximum likelihood, generate data following the true distribution, and estimate the true truncated density.
To establish the convergence, we require \(p_T(\cdot\mid \bm{x})\) to be the unique stationary point of
\(\ell_{\mathrm{obs}}\) over \(\mathcal M\) in order to rule out additional
stationary densities. This condition introduces identifiability for the true density \(p_T\).

\section{Censored Flow Matching for Empirical Data}\label{sec: CFMem}

Let \(\{(Y_i,\Delta_i,\bm X_i):i=1,\ldots,n\}\) be independent and identically distributed copies of \((Y,\Delta,\bm X)\), with \(n\) being the sample size. In this section, we implement censored flow matching based on \(\{(Y_i,\Delta_i,\bm X_i):i=1,\ldots,n\}\).
To estimate the vector field, we parameterize \(v(u,z,\bm x)\) by a fully connected neural network with input \((u,z,\bm x^\top)^\top\) and scalar output. Specifically, we consider
\(
v\in\mathcal{F}_{d+2,1}(l,m),
\)
where \(\mathcal{F}_{d_{\mathrm{in}},d_{\mathrm{out}}}(l,m)\) denotes the class of fully connected neural networks \(f:\mathbb{R}^{d_{\mathrm{in}}}\to\mathbb{R}^{d_{\mathrm{out}}}\) with \(l\) hidden layers and width \(m\). Concretely, any \(f\in\mathcal{F}_{d_{\mathrm{in}},d_{\mathrm{out}}}(l,m)\) admits the representation
\(
f(\bm{s})
=
\bm{W}_{l+1}\,
\sigma\!\Big(
\bm{W}_{l}\,
\sigma\big(
\cdots
\sigma(\bm{W}_1\bm{s}+\bm{b}_1)
\cdots
\big)
+\bm{b}_{l}
\Big)
+\bm{b}_{l+1},
\)
where
\(
\bm{W}_1\in\mathbb{R}^{m\times d_{\mathrm{in}}},
\
\bm{W}_{l+1}\in\mathbb{R}^{d_{\mathrm{out}}\times m},
\
\bm{W}_{\ell}\in\mathbb{R}^{m\times m},\quad \ell=2,\ldots,l,
\)
\(\bm{b}_\ell\) are bias vectors of compatible dimensions, and \(\sigma(\cdot)\) is an elementwise activation function. In this article, we use the ReLU activation function \(\sigma(x)=\max\{0,x\}\).

\subsection{Flow-Based EM Algorithm via Minibatch}\label{sec:ffm}

Let \(v_{\theta^{(k)}}\in \mathcal{F}_{d+2,1}(l,m)\) be the \(k\)th-step vector field. We construct an empirical censored flow-matching loss based on \(\mathcal L_{\mathrm{CFM}}(v_{\theta},q^{(k)})\) using Monte Carlo approximation. Specifically, for each observation \((Y_i,\Delta_i,\bm X_i)\), if \(\Delta_i=1\), the event time is observed and we set
\(
T_i^{(k)}=Y_i.
\)
If \(\Delta_i=0\), the event time is censored and is imputed from the truncated  distribution induced by $v_{\theta^{(k)}}$,
\[
T_i^{(k)}
\sim
q^{(k)}(\cdot\mid \Delta_i=0,Y_i,\bm X_i)
=
p_{T,v_{\theta^{(k)}}}(\cdot\mid \bm X_i,T>Y_i).
\]
This can be accomplished using the pullback-flow procedure described in Theorem~\ref{Theo:for_back}. Under this construction, we sample a minibatch of indices
\(\{i_b\}_{b=1}^B\), where each \(i_b\) is sampled uniformly from \(\{1,\ldots,n\}\) and \(B\) is the batch size. For each \(i_b\), we independently sample
\(
Z_{i_b}\sim p_Z,
\
U_{i_b}\sim \mathrm{Unif}(0,1),
\)
and define
\(
T^{(k)}_{i_b,U_{i_b}}
=
(1-U_{i_b})Z_{i_b}
+
U_{i_b}T^{(k)}_{i_b}.
\)
The minibatch empirical CFM loss is then given by
\[
\widehat{\mathcal L}_{\mathrm{CFM}}^{(k)}(\theta)
=
\frac{1}{B}
\sum_{b=1}^{B}
\left[
v_{\theta}\left(U_{i_b},T^{(k)}_{i_b,U_{i_b}},\bm X_{i_b}\right)
-
\left(T^{(k)}_{i_b}-Z_{i_b}\right)
\right]^2.
\]

Let \(v_\theta\in \mathcal{F}_{d+2,1}(l,m)\). We optimize \(\theta\) using gradient-based optimization based on the \(k\)th-step loss \(\widehat{\mathcal L}_{\mathrm{CFM}}^{(k)}(\theta)\). At each step, we resample a minibatch of indices \(\{i_b\}_{b=1}^B\), construct the minibatch loss at the current \(\theta\), and update \(\theta\) by adaptive moment estimation \citep[Adam;][]{kingma2014adam} using the gradient \(\nabla_\theta \widehat{\mathcal L}_{\mathrm{CFM}}^{(k)}(\theta)\). Repeating these steps yields the \((k+1)\)th-step estimator \(\theta^{(k+1)}\) and corresponding vector field \(v_{\theta^{(k+1)}}\). Iterating over \(k\) yields the final estimate \(\widehat\theta\) and vector field \(v_{\widehat\theta}\). The complete procedure is given in Algorithm~\ref{alg:cfm_minibatch}.

\begin{algorithm}[htbp]
\caption{Censored Flow Matching with Minibatch}
\label{alg:cfm_minibatch}
\begin{algorithmic}[1]
\Require Censored data \(\{(Y_i,\Delta_i,\bm X_i):i=1,\ldots,n\}\); base distribution \(p_Z\); network depth \(l\), width \(m\); batch size \(B\); inner training steps \(O\); outer EM-iterations \(K\).
\Ensure Estimated vector field \( v_{\widehat{\theta}}(u,t,\bm x)\).

\State Parameterize the vector field by \(v_\theta\in \mathcal{F}_{d+2,1}(l,m)\) and initialize \(\theta^{(0)}\).

\For{\(k=0,1,\ldots,K-1\)}
    \State Initialize the inner-loop parameter at \(\theta\leftarrow \theta^{(k)}\).

    \For{\(h=1,2,\ldots,O\)}
        \State Sample a minibatch of indices \(\{i_b\}_{b=1}^B\), where each \(i_b\sim \mathrm{Unif}\{1,\ldots,n\}\).

        \For{\(b=1,\ldots,B\)}
            \State Set \(T_{i_b}^{(k,h)}:=Y_{i_b}\) if \(\Delta_{i_b}=1\); otherwise, compute the pullback value \(z_{Y_{i_b}}^{(k)}\) by solving the backward ODE~\eqref{back_flow} with \(v=v_{\theta^{(k)}}\), \(\bm X=\bm X_{i_b}\), and initial value \(Y_{i_b}\), sample \(Z_{0,i_b}^{(k,h)}\sim p_Z(\cdot\mid Z_0>z_{Y_{i_b}}^{(k)})\), and generate \(T_{i_b}^{(k,h)}\) by solving the forward ODE~\eqref{ode_generation} with \(v=v_{\theta^{(k)}}\), initial value \(Z_{0,i_b}^{(k,h)}\), and covariate \(\bm X_{i_b}\).

            \State Independently draw
            \(
            Z_b^{(h)}\sim p_Z,\ 
            U_b^{(h)}\sim \mathrm{Unif}([0,1])
            \), and construct
            \(
            T_{U_b^{(h)},i_b}^{(k,h)}
            :=
            (1-U_b^{(h)})Z_b^{(h)}
            +
            U_b^{(h)}T_{i_b}^{(k,h)}
            \)
            and
            \(
            V_b^{(k,h)}
            :=
            T_{i_b}^{(k,h)}-Z_b^{(h)}.
            \)
        \EndFor

        \State Compute the minibatch empirical CFM loss:
        \[
        \widehat{\mathcal L}_{\mathrm{CFM},B}^{(k,h)}(\theta)
        :=
        \frac{1}{B}
        \sum_{b=1}^{B}
        \left[
        v_{\theta}\left(
        U_b^{(h)},
        T_{U_b^{(h)},i_b}^{(k,h)},
        \bm X_{i_b}
        \right)
        -
        V_b^{(k,h)}
        \right]^2.
        \]
        \State Update \(\theta\) by one Adam step using
        \(
        \nabla_\theta
        \widehat{\mathcal L}_{\mathrm{CFM},B}^{(k,h)}(\theta).
        \)
    \EndFor

    \State Set \(\theta^{(k+1)}:=\theta\).
\EndFor

\State Set \(\widehat\theta:=\theta^{(K)}\) and output
\(v_{\widehat\theta}(u,t,\bm x).
\)

\end{algorithmic}
\end{algorithm}

In our implementation, we set \(B=500\), \(O=50\), and \(K=50\). The neural network parameters are \(m=100\) and \(l=3\), as three-layer networks provide sufficient flexibility \citep{ismailov2023three}. We apply a log transformation to the observed time, i.e., replacing \(Y_i\) by \(\log Y_i\) in Algorithm~\ref{alg:cfm_minibatch}. Accordingly, we choose \(p_Z\) as the distribution of \(\log E\), where \(E\sim \mathrm{Exp}(1;[0,I])\), \(1\) is the rate parameter, and \(I\) is a prespecified upper bound larger than the maximum observed time. Sampling from \(p_Z(\cdot\mid Z>z)\) is straightforward: for \(z<\log I\), by the memoryless property of the exponential distribution, we sample \(E'\sim \mathrm{Exp}(1;[0,I-\exp(z)])\) and set \(Z=\log\{\exp(z)+E'\}\). The initial parameter \(\theta^{(0)}\) is obtained by applying standard flow matching \citep{liu2022flow,lipman2022flow,lipman2024flow} to the observed pairs \((Y_i,\bm X_i)\), using the same vector-field architecture as in CFM.

\paragraph*{Early Stopping.}
To avoid overfitting, we apply early stopping in the flow-based EM algorithm based on Brier scores \citep{graf1999assessment}. Specifically, we randomly split the data into a training set and a validation set, with \(10\%\) of the observations used for validation. We first estimate the censoring survival function
\(
G(t)=\mathbb P(C>t)
\)
from the training set using the Kaplan--Meier (KM) estimator, denoted by \(\widehat G\). At each outer iteration \(k\), let
\(\widehat S_i^{(k)}(t)\) denote the survival probability at time \(t\) for validation observation \(i\) under the current vector field \(v_{\theta^{(k)}}\), which can be computed via the pullback procedure based on \eqref{pro_equi}. Specifically, letting \(z_{i,t}^{(k)}\) denote the value obtained by transporting \(\log t\) backward to the base space under \(v_{\theta^{(k)}}\) and $\bm{X}_i$, we have
\begin{equation}\label{pull_back}
\widehat S_i^{(k)}(t)
=
\mathbb P\left(Z_0>z_{i,t}^{(k)}\right)
=
\frac{
\exp\left\{-\exp\left(z_{i,t}^{(k)}\right)\right\}
-
\exp\left\{-I\right\}
}{
1-\exp\left\{-I\right\}
}.
\end{equation}
where the last equality follows from the log-exponential base distribution.
We then evaluate
\[
\widehat{\mathrm{BS}}_{\mathrm{val}}^{(k)}(t)
=
\frac{1}{n_{\mathrm{val}}}
\sum_{i\in\mathcal I_{\mathrm{val}}}
\left[
\frac{
\mathbb I(Y_i\le t,\Delta_i=1)
\{\widehat S_i^{(k)}(t)\}^2
}{
\widehat G(Y_i-)
}
+
\frac{
\mathbb I(Y_i>t)
\{1-\widehat S_i^{(k)}(t)\}^2
}{
\widehat G(t)
}
\right],
\]
where \(n_{\mathrm{val}}\) denotes the number of observations in the validation set, \(\mathcal I_{\mathrm{val}}\) denotes the corresponding index set, and \(\widehat G(Y_i-)\) denotes the left limit of the estimated censoring survival function at \(Y_i\).
Accordingly, we compute the validation BS by  integrating
\begin{eqnarray}\label{IBS}
    \widehat{\mathrm{IBS}}_{\mathrm{val}}^{(k)}
=
\frac{1}{\tau_{\mathrm{val}}}
\int_{0}^{\tau_{\mathrm{val}}}
\widehat{\mathrm{BS}}_{\mathrm{val}}^{(k)}(t)\,
\mathrm dt.
\end{eqnarray}
The evaluation interval is fixed across iterations, where its upper endpoint
\(\tau_{\mathrm{val}}\) is restricted by both the support of the estimated censoring distribution and the \(90\%\) quantile of the observed training times. We monitor the minimum validation IBS across outer iterations and stop training if this minimum does not improve for 10 consecutive iterations. 

\subsection{Time-To-Event Generation via CFM}\label{sec: syn_sur}

Generating time-to-event data is useful in many aspects of survival analysis, including event-time imputation for censored observations and synthetic generation when direct data analysis is limited. In the following, we demonstrate that these tasks can be accomplished by CFM.

For the imputation task, suppose we observe a censored subject \((Y^{'},\Delta^{'},\bm X^{'})\) with \(\Delta^{'}=0\), and we aim to impute the latent event time based on $Y^{'}$ and \(\bm X^{'}\). To this end, we apply the pullback-flow sampling procedure in Theorem~\ref{Theo:for_back}: we first compute the pullback value \(\widehat z_{\log(Y^{'})}\) of \(\log(Y^{'})\) via \eqref{back_flow}, with \(v=v_{\widehat{\theta}}\). We then sample
\(
Z_0\sim p_Z(\cdot\mid Z_0>\widehat z_{\log(Y^{'})}),
\)
and solve the learned flow ODE
\begin{eqnarray}\label{syn_dat}
    \frac{\mathrm d Z_u}{\mathrm d u}
=
v_{\widehat\theta}(u,Z_u,\bm X^{'}),
\qquad u\in[0,1].
\end{eqnarray}
The imputed event time is then given by
\(
\widetilde T^{'}=\exp(Z_1).
\)
Note that this procedure provides a nonparametric simulation mechanism for censored observation imputation. By repeatedly generating \(\widetilde T^{'}\), we can obtain summaries such as the mean, median, quantiles, and uncertainty intervals for the censored time.

For the generation task, we need to avoid releasing the original survival dataset
\(\{(Y_i,\Delta_i,\bm X_i);i=1,\ldots,n\}\) for downstream analysis and instead generate a synthetic dataset.
To achieve this, we first construct a data generator for the covariates based on the dataset \(\{\bm X_i;i=1,\ldots,n\}\), for example using standard flow matching \citep{liu2022flow,lipman2022flow,lipman2024flow}. This yields synthetic covariates
\(
\bm X^{\mathrm{syn}}_\ell,\ \ell=1,\ldots,L,
\)
with \(L\) being the desired synthetic sample size. Given each synthetic covariate \(\bm X^{\mathrm{syn}}_\ell\), we generate \(T^{\mathrm{syn}}_\ell\) using the same event-time generation procedure in \eqref{syn_dat}, with \(\bm X^{'}=\bm X^{\mathrm{syn}}_\ell\) and \(Z_0\sim p_Z\) without truncation. 
Repeating this procedure for \(\ell=1,\ldots,L\) gives a synthetic time-to-event dataset
\(
\{(T^{\mathrm{syn}}_\ell,\bm X^{\mathrm{syn}}_\ell):\ell=1,\ldots,L\}
\)
 for downstream analysis.

\paragraph*{Empirical Privacy Evaluation.}
One objective of synthetic event-time generation is to protect the privacy in the original dataset. We therefore evaluate privacy by comparing the original censored event-time
with the synthetic complete event-time data.
A challenge is that the original and synthetic datasets have different observation structures. Moreover, the privacy risk of interest concerns the latent uncensored event time rather than the observed censoring time; that is, we evaluate whether the synthetic data reveal information about an individual's true event time \(T\) beyond that contained in the observed censored data \(Y\). Thus, standard privacy measures, such as nearest-neighbor-based metrics \citep{yale2020generation} and membership inference attacks \citep{zhang2022membership}, cannot be directly applied without accounting for censoring.

To address this issue, we propose a privacy evaluation comparing synthetic and original event times. Let \(G(t)=\mathbb P(C\geq t)\) denote the censoring survival function, estimated by the KM estimator from the original data \((Y_i,\Delta_i)\). We define the censoring-adjusted nearest-neighbor distance for the training data as
\[
\widehat D_{\mathrm{train}}
=
\frac{1}{n}\sum_{i=1}^n
\frac{\Delta_i}{G(Y_i-)}
\min_{1\leq\ell\leq L}
\left|T_\ell^{\mathrm{syn}}-Y_i\right|,
\]
where \(\Delta_i/G(Y_i-)\) is an inverse probability of censoring weight accounting for right censoring. Under some conditions, \(\widehat D_{\mathrm{train}}\) is an unbiased estimator of the distance
\(
D_{\mathrm{train}}
=
\mathbb E\left[
\frac{1}{n}\sum_{i=1}^n
\min_{1\leq\ell\leq L}
|T_\ell^{\mathrm{syn}}-T_i|
\right];
\)
see Part~\ref{sec:privacy-proof} of the Supplementary Material. Thus, \(\widehat D_{\mathrm{train}}\) quantifies the proximity between synthetic event times and latent event times in the training data. Given an independent testing dataset, we similarly define \(\widehat D_{\mathrm{test}}\) by replacing the training observations with testing observations. Comparing \(\widehat D_{\mathrm{train}}\) and \(\widehat D_{\mathrm{test}}\) assesses whether the synthetic data are excessively close to training individuals, with a substantially smaller \(\widehat D_{\mathrm{train}}\) indicating potential memorization and privacy risk.

\subsection{Survival Density Estimation via CFM}
CFM can also provide a truncated density estimator for the survival time via \eqref{densi}.
Note that CFM is trained on the log-time scale. Let \(R=\log T\). Given a covariate value \(\bm X=\bm x\) and a truncation time \(Y>0\), the truncated density of \(R\) can be estimated according to \eqref{densi} as
\[
\log \widehat p_R(r\mid \bm X=\bm x,R\geq \log Y)
=
\log p_Z(\widehat z_0)
-
\log \mathbb P_Z(Z_0\geq \widehat z_{\log Y})
-
\int_0^1
\partial_z v_{\widehat\theta}(u,\widehat z_u,\bm x)\,\mathrm{d}u,
\ r\geq \log Y,
\]
where \(r=\log t\); \(\widehat z_0\) is the pullback of \(r\) under the inverse flow induced by \(v_{\widehat\theta}\), and \(\{\widehat z_u;u\in [0,1]\}\) denotes the corresponding forward trajectory from \(\widehat z_0\) to \(r\), satisfying
\(
    \frac{\mathrm d \widehat z_u}{\mathrm d u}
    =
    v_{\widehat\theta}(u,\widehat z_u,\bm x);
\)
\(\widehat z_{\log Y}\) is the pullback of \(\log Y\) under the same inverse flow.
The truncated density of \(T\) is then estimated by
\begin{eqnarray}\label{truncted}
&&\log \widehat p_T(t\mid \bm X=\bm x,T\geq Y)\nonumber\\
&=&
\log \widehat p_R(\log t\mid \bm X=\bm x,R\geq \log Y)
-
\log t\nonumber\\
&=&
\log p_Z(\widehat z_0)
-
\log \mathbb P_Z(Z_0\geq \widehat z_{\log Y})
-
\int_0^1
\partial_z v_{\widehat\theta}(u,\widehat z_u,\bm x)\,\mathrm{d}u
-
\log t,
\qquad t\geq Y.
\end{eqnarray}

Compared with hazard-based methods, the above density provides a direct distributional characterization of patient-specific event times. In particular, it allows us to evaluate the full truncated conditional density \(p_T(t\mid \bm X=\bm x,T\geq Y)\) for a censored patient, rather than relying on hazard scores or survival probabilities as indirect summaries of the event-time density.

\section{Simulation}\label{sec:sim}

In this section, we assess the performance of CFM in generating latent time-to-event data from censored observations. For each simulation setting, we independently sample covariates
\(
\bm{X}_i \sim N(\mathbf 0,\bm{I}_d),\ i=1,\ldots,n
\), with $d = 10$. Given \(\bm{X}_i\), the event time \(T_i\) is generated from one of several data-generating models described below, where the generation is implemented such that \(T_i\in[0,1]\) holds with high probability. 
To mimic right censoring, we generate censoring times from a truncated exponential distribution on \([0,1]\):
\(
C_i \sim \mathrm{Exp}\{\lambda_C\exp(\bm\beta_C^\top \bm X_i); [0, 1]\},
\)
where the entries of \(\bm\beta_C\in \mathbb{R}^d\) are independently generated from \(N(0,0.1^2)\). The observed time and censoring indicator are then defined as
\(
Y_i=\min(T_i,C_i),\
\Delta_i=\mathbb{I}(T_i\leq C_i).
\)
To obtain a suitable censoring level, we choose \(\lambda_C\) to yield approximately \(50\%\) censoring.

The generation of \(T_i\) follows four different scenarios listed below:
\begin{itemize}
   \item \textbf{Linear CoxPH Setting:} We generate \(T_i\) via a linear Cox proportional hazards model:
\begin{eqnarray}\label{CoxPH}
    h(t\mid \bm X_i)=h_0(t)\exp(\bm X_i^\top\bm\beta),
\end{eqnarray}
where \(\bm\beta\in\mathbb R^d\) is sampled with independent entries from \(N(0,0.1^2)\). The baseline hazard is generated using a Fourier basis expansion,
\(
h_0(t)=\exp\{\bm F(t)^\top\bm\gamma\},
\)
where
\(
\bm F(t)=
\bigl(1,\sin(2\pi t),\cos(2\pi t),\ldots,
\sin(2\pi K_Ft),\cos(2\pi K_Ft)\bigr)^\top
\)
with \(K_F=3\), and the coefficients \(\bm\gamma\in\mathbb R^{1+2K_F}\) are generated with independent entries from \(N(0,1)\). Given \(\bm X_i\), the event time \(T_i\) is generated by inverse transform sampling from the cumulative hazard $H(t\mid \bm X_i)=\int_0^t h(s\mid \bm X_i)\,\mathrm{d}s$, following the procedure in \eqref{remark_hazard}.
We then rescale \(T_i\) as \(T_i/a\), \(i=1,\ldots,n\), where \(a\) is chosen so that the rescaled event times lie in \([0,1]\) with probability greater than \(95\%\).

   \item \textbf{Linear-AFT Setting:} We generate \(T_i\) from an accelerated failure time model:
\begin{eqnarray}\label{eqa:AFT}
    \log T_i=\bm X_i^\top\bm\beta+\epsilon_i,
\qquad
\epsilon_i\sim N(\mu_\epsilon,\sigma_0^2).
\end{eqnarray}
The regression coefficient \(\bm\beta\in\mathbb R^d\) is generated with independent entries from \(N(0,(1/6)^2)\), and we set
\(
\sigma_0^2=\operatorname{Var}(\bm X_i^\top\bm\beta)/\mathrm{SNR},
\)
where the signal-to-noise ratio (\(\mathrm{SNR}\)) is fixed at \(10\) in the simulation. The location parameter \(\mu_\epsilon\) is chosen so that the generated event times lie in \([0,1]\) with probability greater than \(95\%\).

   \item \textbf{Nonlinear Setting 1:} We generate \(T_i\) through a nonlinear transformation model
\[
T_i=\operatorname{sigmoid}\{m(\bm X_i)+\epsilon_i\},
\qquad
\epsilon_i\sim N(0,\sigma^2),
\]
where
\(
m(\bm X_i)
=
2\left[
\sin(\eta_{i1})
+\tanh(1-\eta_{i2}^2)
+\tanh(\eta_{i3})
+\allowbreak \{\cos(\eta_{i4}) -1\}
\right],
\)
with \(\eta_{ik}=\bm X_i^\top\bm\beta_k\), \(k=1,\ldots,4\), and \(
\operatorname{sigmoid}(u)=\frac{1}{1+\exp(-u)}
\). The vectors \(\bm\beta_1,\ldots,\bm\beta_4\in\mathbb R^d\) are independently generated with entries from \(N(0,1)\). We set
\(
\sigma^2=\frac{\operatorname{Var}\{m(\bm X_i)\}}{\mathrm{SNR}},
\)
with \(\mathrm{SNR}\) fixed at \(10\).

  \item \textbf{Nonlinear Setting 2:} We generate \(T_i\) from a nonlinear mixture model. First, a latent class indicator is generated as
\(
B_i\sim \operatorname{Bernoulli}(0.5),
\)
independently of \(\bm X_i\). Then
\[
T_i=\operatorname{sigmoid}\{m_{B_i}(\bm X_i)+\epsilon_i\},
\qquad
\epsilon_i\sim N(0,\sigma^2),
\]
where \(m_0(\bm X_i)\) is the same nonlinear function as in Nonlinear setting 1,
and \(m_1(\bm X_i)\) is given by
\(
2\left[
-\sin^2(\eta_{i1})
+\tanh(\eta_{i2})
-\tanh(\eta_{i3})
+\sin(\eta_{i4})
\right].
\)
Here \(\eta_{ik}=\bm X_i^\top\bm\beta_k\), \(k=1,\ldots,4\), and \(\bm\beta_1,\ldots,\bm\beta_4\in\mathbb R^d\) are independently generated with entries from \(N(0,1)\). We set
\(
\sigma^2=\frac{\operatorname{Var}\{m_{B_i}(\bm X_i)\}}{\mathrm{SNR}}
\)
with \(\mathrm{SNR}\) fixed at \(10\).
\end{itemize}

\paragraph*{Comparison Methods.}
We compare CFM with several representative survival modeling and generative baselines. The first is an uncensored flow-matching model \citep{liu2022flow,lipman2022flow,lipman2024flow} trained directly on the observed censored outcomes without censoring adjustment, denoted as FM. We also consider survival-likelihood-based generative baselines, which first estimate the conditional hazard function using survival likelihoods and then generate event times from the estimated hazard; similar approaches have been proposed in \citet{norcliffe2023survivalgan}.

To estimate hazard functions, we consider the standard Cox proportional hazards model fitted by partial likelihood \citep{cox1975partial} and full likelihood \citep{royston2002flexible}, denoted as CoxPH-P and CoxPH-F, respectively. Both use model~\eqref{CoxPH}, assuming a linear covariate effect with a flexible baseline hazard. For CoxPH-P, the regression coefficient is estimated by partial likelihood and the baseline cumulative hazard by the Breslow estimator \citep{cox1975partial}. For CoxPH-F, the baseline hazard is modeled using splines and estimated jointly with the regression coefficient by full likelihood.

We further consider several nonparametric hazard estimators. Random survival forest \citep{ishwaran2008random}, denoted as RSF, is a nonparametric ensemble method for survival estimation within terminal nodes. CoxTime \citep{kvamme2019time} is a neural extension of the Cox partial likelihood approach with time-varying covariate effects, using
\(
h(t\mid \bm X_i)=h_0(t)\exp\{g_\theta(t,\bm X_i)\},
\)
where \(g_\theta(t,\bm X_i)\) is a neural network. We also consider a full-likelihood neural hazard model \citep{su2026nonparametric}, denoted as Hazard-Neural, which directly models
\(
h(t\mid \bm X_i)=\exp\{g_\theta(t,\bm X_i)\}.
\)

Furthermore, we consider parametrized methods for time-to-event generation. We include the accelerated failure time model~\eqref{eqa:AFT}, denoted as AFT, fitted by maximum likelihood under a log-normal error distribution \citep{kalbfleisch2002statistical}. We also consider VAE-based survival generation \citep{xiu2020variational,apellaniz2024leveraging}, as described in Part~\ref{remark_VAE} of the Supplementary Material. The VAE follows a similar philosophy to CFM: both generate event times by transporting a latent random variable to the target distribution. However, VAEs typically rely on a direct transport from latent variables to event times, where the distribution of learned latent representations may not match the prespecified prior used for generation. In contrast, CFM learns a continuous, invertible ODE-based transport, avoiding potential latent distribution mismatch while providing greater modeling flexibility through its continuous dynamics.

For the forest-based method, we set the number of trees to
\(500\) and the terminal node size to \(15\). For neural-network-based and VAE methods, we use a multilayer perceptron with three hidden layers, each
containing \(100\) hidden units. Early stopping is employed to avoid
overfitting. The estimated hazard functions are then used to generate
time-to-event data by inverse-transform sampling from the estimated cumulative
hazard function, as in \eqref{remark_hazard}.

\paragraph*{Comparison Measure.}
We evaluate generated event times using the average conditional Wasserstein distance
\[
\mathrm{Err}
=
\mathbb E_{\bm X}
\left[
W_2\left(
\widetilde T,T\mid \bm X,\ 0<T,\widetilde T<1
\right)
\right],
\]
where \(\widetilde T\) and \(T\) denote the generated and true event times, respectively. Here, \(W_2(\widetilde T,T\mid \bm X,\ 0<T,\widetilde T<1)\) measures the second-order Wasserstein distance between their conditional distributions truncated to \((0,1)\). In implementation, we generate \(500\) test covariate values \(\{\bm X_i\}_{i=1}^{500}\), and for each \(\bm X_i\), generate \(50\) true and \(50\) synthetic event times from the corresponding conditional distributions. We compute the Wasserstein distance for each \(\bm X_i\) and average over the \(500\) covariate values to approximate \(\mathrm{Err}\).

\begin{figure}[htbp]
    \centering
    \includegraphics[width=0.9\linewidth]{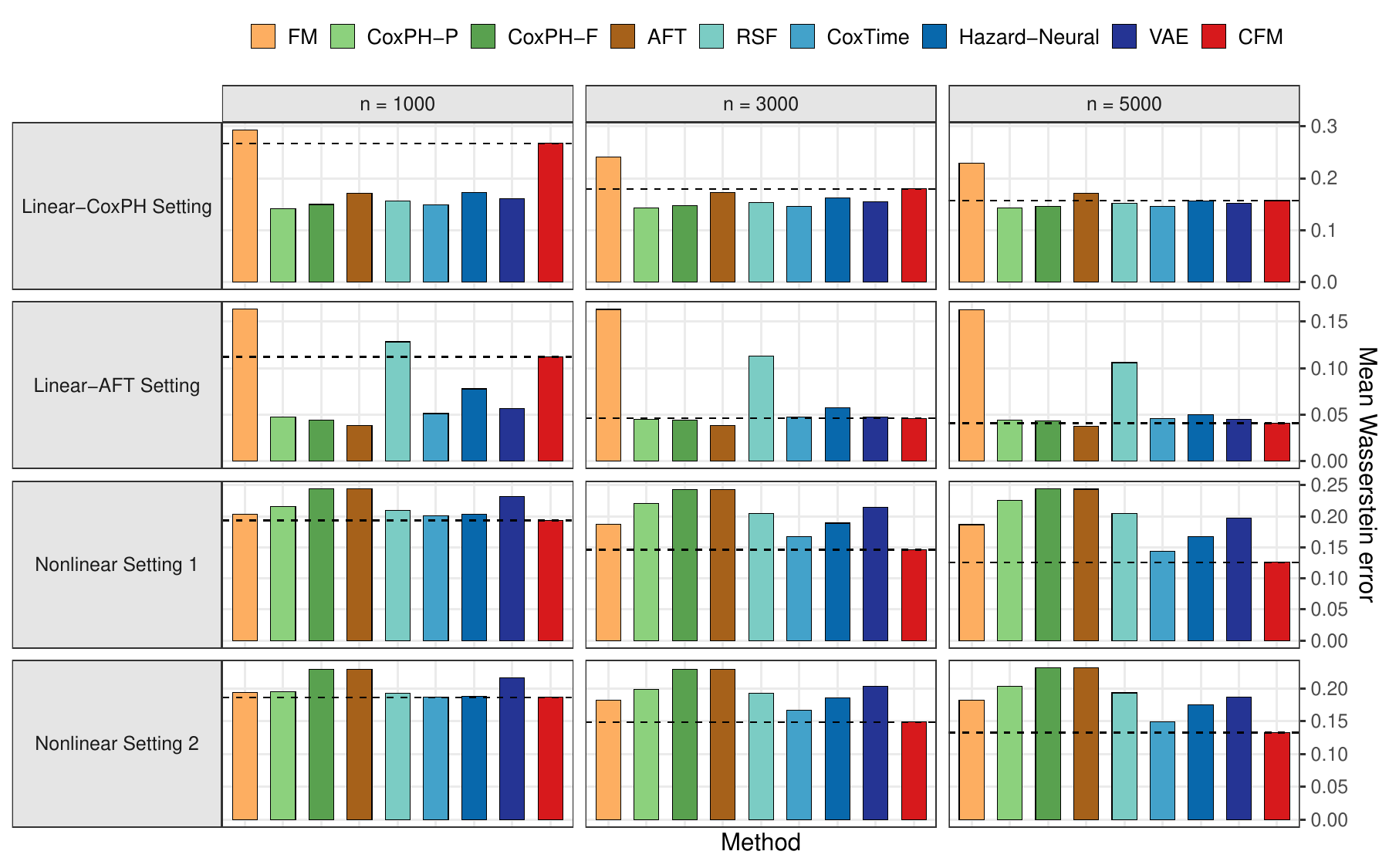}
    \caption{Mean averaged conditional Wasserstein errors over 100 simulation replications across simulation settings and sample sizes.}
    \label{fig:simulation-error}
\end{figure}

To examine censoring recovery, we also consider a truncated error. Let \(Q_T(u\mid \bm X,0<T<1)\) and \(Q_{\widetilde T}(u\mid \bm X,0<\widetilde T<1)\) denote the conditional quantile functions of the true and generated event times for \(u\in[0,1]\). For a truncation level \(q\in[0,1)\), define
\[
\mathrm{Err}(q)
=
\mathbb E_{\bm X}
\left[
\left\{
\frac{1}{1-q}
\int_q^1
\left[
Q_{\widetilde T}(u\mid \bm X,0<\widetilde T<1)
-
Q_T(u\mid \bm X,0<T<1)
\right]^2
\,\mathrm{d}u
\right\}^{1/2}
\right].
\]
When \(q=0\), \(\mathrm{Err}(q)\) reduces to the original conditional Wasserstein error, while larger \(q\) focuses on the upper tail of the conditional event-time distribution. In implementation, \(\mathrm{Err}(q)\) uses the same sampling scheme as \(\mathrm{Err}\), after discarding the lowest \(q\)-fraction of the sorted true and generated samples for each \(\bm X_i\).

\paragraph*{Simulation Results.}
We implement 100 simulation replications for each setting. The Wasserstein distances under different sample sizes, methods, and settings are reported in Figure~\ref{fig:simulation-error}. Our results show that, in the linear settings, the best generation methods are the corresponding oracle-type models, i.e., CoxPH for the proportional-hazards setting and AFT for the accelerated-failure-time setting, respectively. Our method is inferior to some methods in the linear settings when the sample size is small (\(n=1000\)), but its performance improves more significantly as the sample size increases.
Note that our method consistently outperforms FM, as it addresses the censoring issue in time generation during vector field training.

\begin{figure}[htbp]
    \centering
    \includegraphics[width=0.9\linewidth]{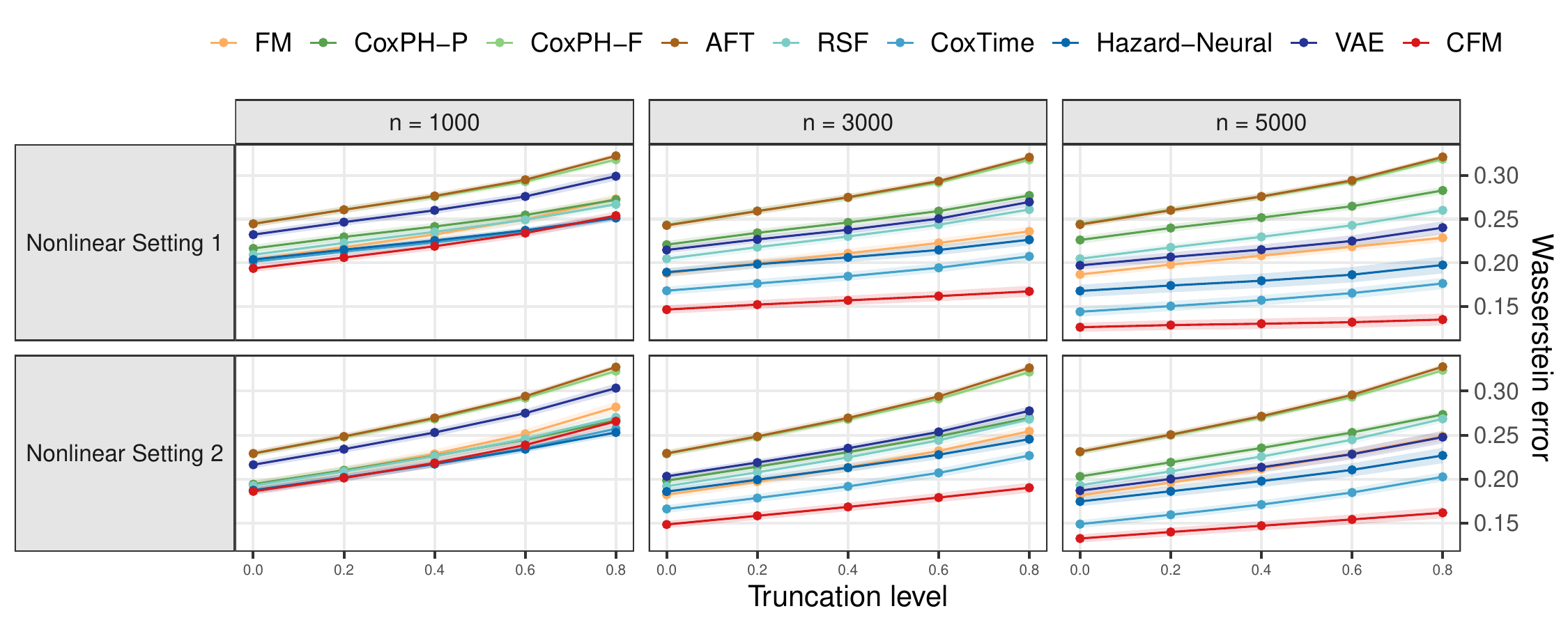}
    \caption{Mean truncated conditional Wasserstein errors across truncation levels for the nonlinear settings, with 95\% confidence intervals computed from 100 simulation replications.}
    \label{fig:simulation-truncation-error}
\end{figure}

For the nonlinear cases, we find that our method consistently outperforms the competing methods, and the improvement becomes more significant as the sample size increases. For the nonlinear cases, we also examine the truncated error \(\mathrm{Err}(q)\) in Figure~\ref{fig:simulation-truncation-error}. We similarly find that our method performs satisfactorily, especially when the sample size is large and the truncation level is high. 
These results suggest the stronger expressive power of CFM in generating event times from large datasets. 
This ability stems from fitting a continuous generative path in CFM, which goes beyond the generation mechanism of hazard-based and VAE-based methods.

\section{Real Data Analysis}\label{sec:data}

In this section, we apply CFM to synthetic generation and predictive analysis for time-to-event data using database from the Organ Procurement and Transplantation Network (OPTN) and the Medical Information Mart for Intensive Care IV (MIMIC-IV)  \citep{johnson2020mimic}.

\paragraph*{Synthetic Survival Analysis on OPTN Database.}
We evaluate CFM for synthetic data analysis using de-identified data extracted from the OPTN database, which contains records of transplant events performed in the United States and reported to OPTN. 
This dataset contains recipient-level records for individuals who underwent a first deceased-donor kidney transplant between January 1, 2015, and December 31, 2018.
For the analysis, we focus on recipients aged 18 years or older and consider the transplantation date of each recipient as the baseline time. 
The event of interest is the first occurrence of either all-cause recipient death or kidney graft failure within one year ($365$ days) after transplantation. 
Recipients who do not experience the event are censored at the earlier of their last follow-up date and $365$ days after transplantation. 
The resulting analytic cohort consists of $45738$ recipients, including $2754$ events and $42984$ censored observations, corresponding to an overall censoring proportion of $93.98\%$.
For each recipient, we collect covariates including age, recorded gender, race/ethnicity, and body mass index. 
Table~\ref{tab:srtr-summary-full} in Part~\ref{sec:covariate} of the Supplementary Materials gives the summary statistics of the analytic cohort.

For data analysis, we standardize the continuous covariates in the dataset to have zero mean and unit variance, while the categorical covariates are encoded using one-hot encoding. After encoding, the resulting covariate vector has dimension $d = 7$.
We then apply CFM to generate a synthetic survival dataset based on the OPTN database.

For implementation, we perform an event-stratified split of the analytic cohort into training and testing sets at an approximately $80/20$ ratio, yielding $36421$ training and $9317$ testing subjects. The stratified splitting strategy is used to preserve approximately the same censoring proportion in the two sets. We then learn a covariate generator using standard flow matching \citep{liu2022flow}, together with a conditional time-to-event generator using the proposed CFM, based on the training set. Accordingly, we generate new covariates from the learned covariate flow model and then generate the corresponding event times conditional on these generated covariates using CFM. In this way, we obtain a synthetic dataset of the form $(\widetilde{\bm X},\widetilde T)$ from the original observations of the form $(\bm X,Y,\Delta)$.

\begin{figure}[ht!]
    \centering
    \includegraphics[width=0.65\linewidth]{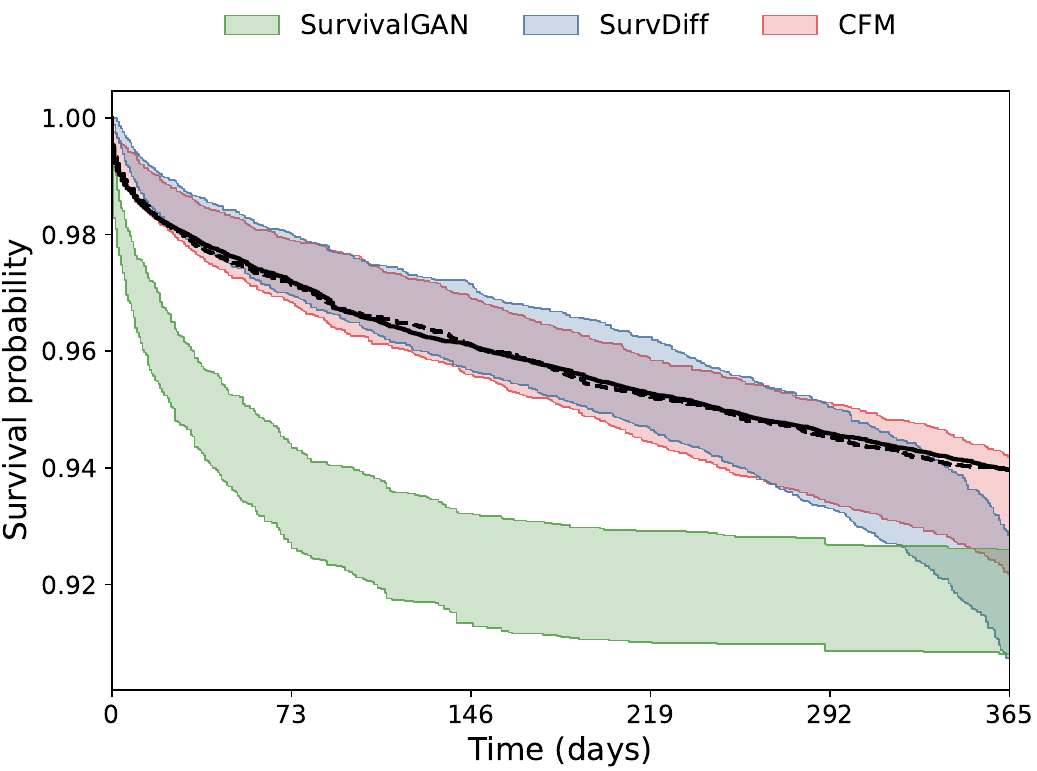}
    \caption{Survival curves estimated using the KM method for the real and synthetic data. The solid and dashed black curves represent the survival curves estimated from the real training and testing sets, respectively. For each synthetic method, the shaded band represents the pointwise minimum and maximum of the KM estimates across 200 independently generated synthetic datasets.
    }
\label{fig:km_curves_training_testing}
\end{figure}

We compare CFM with two baselines: SurvivalGAN \citep{norcliffe2023survivalgan} and SurvDiff \citep{brockschmidt2025survdiff}. Both methods are trained on the training set and aim to generate synthetic datasets of the form \((\widetilde{\bm X},\widetilde Y,\widetilde\Delta)\); see Part~\ref{sec:tuning_setting} of the Supplementary Material for their tuning settings.
For each of the three fitted generative models, we generate $200$ replicate synthetic datasets, each consisting of $5000$ observations. Figure~\ref{fig:km_curves_training_testing} presents the estimated survival curves obtained from the synthetic datasets generated by each method, together with those estimated from the real training and testing sets.

\begin{figure}[!t]
    \centering\hspace{-0.06\linewidth}
    \includegraphics[width=0.8\linewidth]{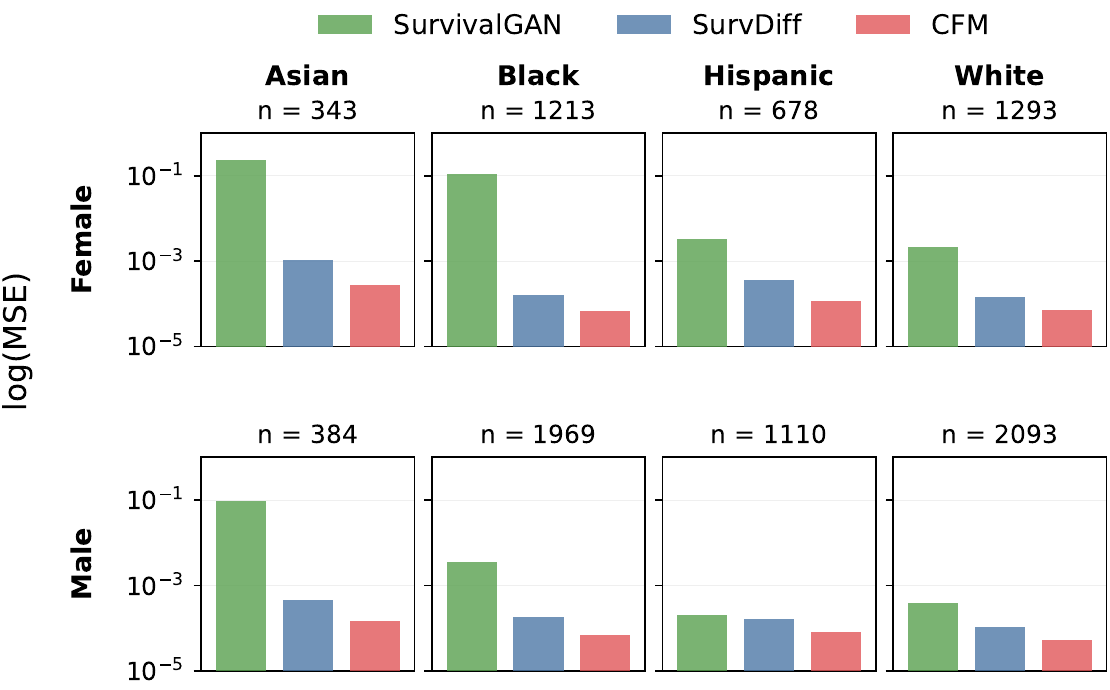}
    \caption{Mean squared error (MSE) between the synthetic and real KM curves within subgroups jointly defined by race/ethnicity and recorded gender. For each method, 200 synthetic datasets, each containing 5000 samples, are independently generated. Within each subgroup, we compute the average MSE across the 200 synthetic datasets, where, for each synthetic dataset,
\(
\mathrm{MSE}
=
\frac{1}{500}\sum_{j=1}^{500}
\{
\widehat{S}^{\mathrm{syn}}(t_j)
-
\widehat{S}^{\mathrm{real}}(t_j)
\}^{2}.
\)
Here, \(\widehat{S}^{\mathrm{syn}}(t_j)\) denotes the KM estimate at time \(t_j\) within the corresponding subgroup, \(\widehat{S}^{\mathrm{real}}(t_j)\) denotes the corresponding KM estimate from the real testing data within the same subgroup, and \(t_j\) is defined on a uniform grid of 500 time points over \([0,365]\) days.}
    \label{fig:km_race_gender_mse_testing}
\end{figure}

{As shown in Figure~\ref{fig:km_curves_training_testing}, the survival probability estimated from SurvivalGAN decreases sharply early in the follow-up period and remains below the corresponding real-data curves. SurvDiff and CFM produce survival bands that more closely resemble those estimated from the real data. However, the SurvDiff curve deviates more substantially from the real-data curves in the tail of the survival distribution, whereas CFM remains closer throughout the follow-up period.}

{
These discrepancies may reflect the challenge of generating observations of the form $(\widetilde{\bm X},\widetilde Y,\widetilde\Delta)$ as in SurvivalGAN and SurvDiff, which requires jointly accounting for uncertainty in both the event-time distribution and the censoring mechanism. 
Given the high censoring proportion in our dataset ($93.98\%$), uncertainty associated with censoring may dominate the synthetic data-generation process, especially in the tail of the survival curve where only a small number of synthetic individuals remain under observation. This may introduce greater instability in the tail of the survival distribution, with the synthetic survival curves showing larger deviations from the real-data curves and a stronger tendency to underestimate the survival probability.
In contrast, CFM directly generates latent event times and therefore does not introduce additional source of censoring uncertainty into the data-generation process. The resulting synthetic datasets yield survival curves that most closely track those estimated from the real data.
}

{
To evaluate subgroup-level generation accuracy, we further quantify the differences between the survival curves estimated from the synthetic and testing data across covariate-defined subpopulations. Figure~\ref{fig:km_race_gender_mse_testing} presents the subgroup-specific differences for subpopulations jointly stratified by race/ethnicity and recorded gender, while Figure~\ref{fig:km_subgroup_mse_testing} in the Supplementary Material presents additional comparisons across body mass index (BMI) subgroups. Across these subgroups, CFM reproduces the survival curves estimated from the testing data more closely than the competing methods.
}

Furthermore, we compute the censoring-adjusted nearest-neighbor distances \(\widehat D_{\mathrm{train}}\) and \(\widehat D_{\mathrm{test}}\) between the synthetic data and the training and testing sets, respectively. We report the overall difference \(\widehat D_{\mathrm{train}}-\widehat D_{\mathrm{test}}\), as well as subgroup-specific differences for the subgroups in Figure~\ref{fig:km_race_gender_mse_testing}. To this end, we randomly split the data into training and testing sets and generate synthetic data using CFM as before, and repeat this procedure 50 times. Table~\ref{tab:cfm-privacy-race-gender} reports the average difference \(\widehat D_{\mathrm{train}}-\widehat D_{\mathrm{test}}\) and its confidence interval for the overall population and each subgroup. We observe that the differences are close to zero, with the corresponding confidence intervals covering zero. These results suggest that the synthetic event times do not exhibit evidence of memorizing the training data during synthetic data generation, for both the global population and subpopulations.

\begin{table}[htbp]
\centering
\caption{Privacy evaluation for CFM based on censoring-adjusted nearest-neighbor distances. Each entry reports the average of \(\widehat D_{\mathrm{train}}-\widehat D_{\mathrm{test}}\) and its 95\% confidence interval, computed across 50 replications.}
\label{tab:cfm-privacy-race-gender}
\resizebox{0.95\linewidth}{!}{\begin{tabular}{lcccc}
\toprule
\textbf{Gender} & \textbf{Asian} & \textbf{Black} & \textbf{Hispanic} & \textbf{White} \\
\midrule
Female & $-0.004\;(-0.077, 0.070)$ & $0.001\;(-0.006, 0.008)$ & $-0.007\;(-0.023, 0.009)$ & $0.002\;(-0.004, 0.008)$ \\
\addlinespace
Male & $-0.029\;(-0.061, 0.003)$ & $0.001\;(-0.002, 0.004)$ & $-0.008\;(-0.024, 0.008)$ & $-0.002\;(-0.004, 0.001)$ \\
\midrule
\textbf{Overall} & \multicolumn{4}{c}{$9.2\times 10^{-5}\;(-0.0002, 0.0004)$} \\
\bottomrule
\end{tabular}
}
\end{table}

Overall, CFM maintains comparatively high fidelity in capturing survival distributions across different covariate-defined subpopulations, while exhibiting favorable empirical privacy characteristics. This balance between data fidelity and privacy showcases the potential of CFM-generated synthetic data to serve as a surrogate for the original data in downstream survival analyses.

\paragraph*{Survival Prediction on MIMIC-IV Database.}
We apply CFM to predict patient survival times from Intensive Care IV (MIMIC-IV) database \citep{johnson2020mimic}. 
For each patient in the database, we retain the first recorded intensive care unit (ICU) stay and use the corresponding ICU admission time as the baseline. 
To exclude extremely short ICU stays, we restrict the analysis to patients who remained in the ICU for at least four hours. 
After data preprocessing, the resulting cohort consists of 64575 patients.
The event of interest is all-cause mortality, and the event time is measured from ICU admission to death. 
For patients without an observed death, follow-up is administratively censored one year after discharge from the hospitalization \citep{johnson2020mimic}. 
Accordingly, the observed survival time is the time from ICU admission to either death or the administrative censoring time, whichever occurs first, together with an indicator recording whether death is observed. 
The resulting cohort has an overall censoring rate of 74.79\%.

We consider a collection of demographic, admission, and clinical characteristics as baseline covariates, including age at ICU admission, gender, race, insurance status, admission type, marital status, and the type of ICU at admission. 
We additionally include indicators for six major comorbidity categories: diabetes, heart failure, renal disease, respiratory disease, cancer, and sepsis. 
For categorical covariates, levels represented by fewer than 20 patients are combined into an ``Other'' category and subsequently encoded using one-hot encoding. 
Age is standardized to have zero mean and unit variance, while the binary and one-hot encoded variables are retained as \(0\)--\(1\) indicators. 
After encoding, the resulting covariate vector has dimension \(d=68\).

\begin{figure}[t]
    \centering
    \includegraphics[width=\textwidth]{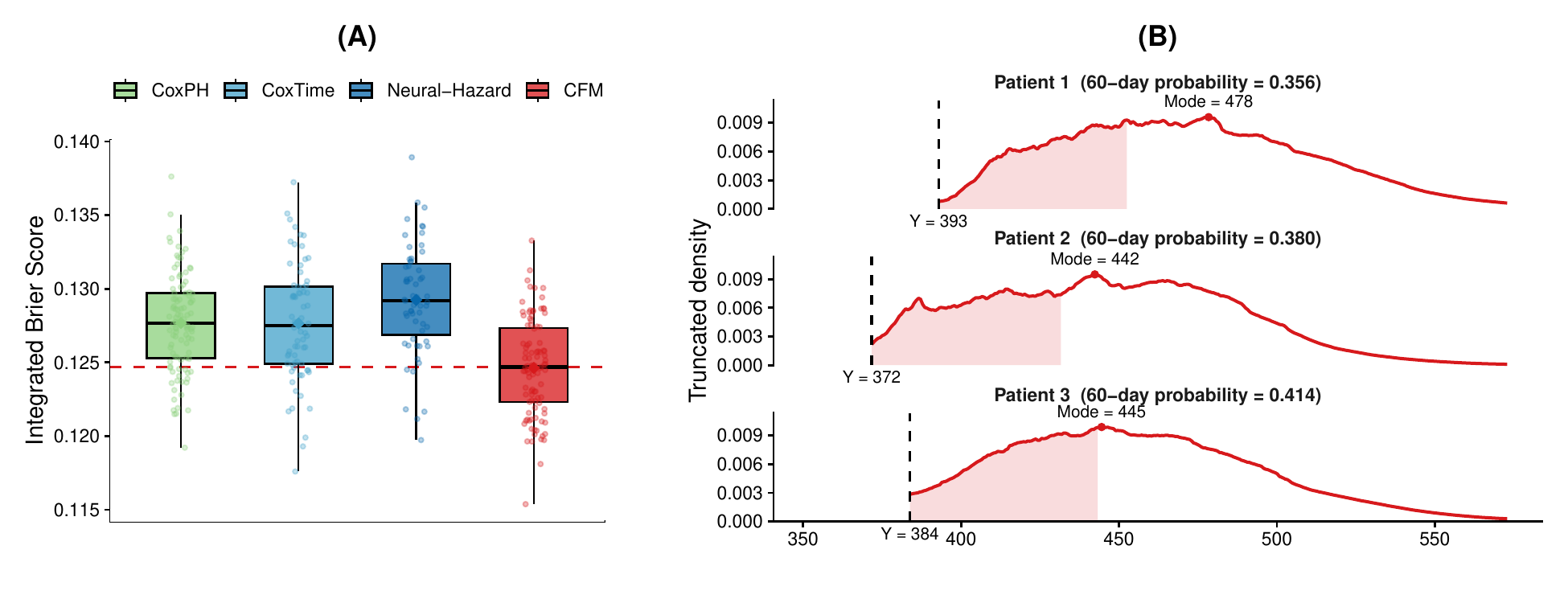}
    \caption{
    Analysis of the MIMIC-IV survival data.
    \textbf{(A)} Boxplots of the test IBS over 100 random train--test splits for CoxPH, CoxTime, Hazard-Neural, and CFM. The dashed red line indicates the median IBS of CFM.
    \textbf{(B)} Estimated patient-specific truncated event-time densities for three representative censored patients. For each patient, the dashed vertical line indicates the observed censoring time \(Y\) in days, the shaded region represents the estimated probability mass over the subsequent 60 days, and the red point indicates the mode of the estimated truncated density.
    }
    \label{fig:mimic}
\end{figure}

We compare CFM with three hazard-based methods considered in the simulation studies, including CoxPH, CoxTime, and Hazard-Neural. 
For evaluation, we repeatedly and randomly split the cohort into training and test sets, with 30000 patients used for training and 3000 patients used for testing in each replication. 
We repeat this procedure 100 times and fit all competing methods using the same training and test splits. 
The prediction performance is evaluated using the IBS (see \eqref{IBS}) on the test data. The computation of IBSs for different methods is given in Part~\ref{sec: IBS} of the Supplementary Material, with lower values indicating better predictive accuracy.
All methods are implemented using the same settings as those described in the simulation studies.
The resulting prediction results are presented in Figure~\ref{fig:mimic}(A). 
Overall, we find that CFM achieves competitive predictive performance, with lower IBS values than other methods.

To illustrate the ability of CFM to recover latent event-time distributions, we fit CFM to the full cohort and illustrate the estimated truncated event-time densities
\(
\widehat p_T(t\mid \bm X,T>Y)
\)
obtained via~\eqref{truncted} for three censored patients, as shown in Figure~\ref{fig:mimic}(B).
We observe that, after the observed follow-up time \(Y\), the estimated density generally increases and then decreases, yielding a high-density region that indicates when the latent event time is most likely to occur conditional on survival beyond \(Y\).
The location and concentration of this high-density region vary substantially across patients, reflecting patient-specific differences in the timing and uncertainty of the latent event time.
These results provide a direct illustration of patient-specific event-time distributions conditional on the observed censoring information, allowing us to characterize periods of elevated event risk for patients and potentially inform the timing of subsequent monitoring and follow-up.

\section{Discussion}\label{sec:ids}

In this article, we introduce Censored Flow Matching (CFM), a continuous generative framework for learning conditional time-to-event distributions from censored survival data. 
CFM models the event-time distribution through an ODE-based optimal transport, while incorporating censoring information through a novel flow-based EM procedure built upon a pullback-flow step. 
Beyond conventional survival prediction, this procedure provides direct access to the full latent event-time distribution, enabling synthetic event-time generation given covariates, as well as patient-specific truncated generation and density estimation conditional on an observed censoring time. 
We establish theoretical guarantees for the proposed procedure and demonstrate its performance through simulation studies. Applications to the OPTN and MIMIC-IV databases further show that CFM enables accurate synthetic time-to-event generation, competitive survival prediction, and flexible characterization of latent event-time distributions.

CFM is closely related to recent diffusion-based EM methods \citep{rozet2024learning,zhang2025diffputer}, both of which alternate between completing latent components and updating continuous generative models. Their key distinction lies in how the E-step is carried out for sampling latent data. Unlike diffusion-based methods, CFM is specifically designed for censoring-constrained latent variables through the monotonicity of ODE flows, offering a theoretically grounded E-step  for censored generative modeling.
More generally, our approach can be extended beyond the right-censoring setting considered here to left and interval censoring, which can be accommodated by incorporating the corresponding lower and/or upper bounds through the same monotonicity property.

The current framework has limitations. First, CFM considers only static covariates and does not incorporate time-varying information in learning the flow.
Second, the proposed nearest-neighbor privacy measure is intended for empirical evaluation and does not aim for a formal privacy guarantee for time-to-event. It would therefore be of interest to extend CFM to accommodate time-varying covariates and to develop theoretically grounded privacy evaluations for synthetic survival data. Such developments would enhance the flexibility and privacy safeguards of the proposed framework, broadening its applicability to longitudinal and privacy-sensitive settings.

\section*{Acknowledgement}
J. T and A. Z. were supported in part by NIH Grant R01HL169347. 
This work was supported in part by Health Resources and Services Administration contract 234-2005-370011C. The content is the responsibility of the authors alone and does not necessarily reflect the views or policies of the Department of Health and Human Services, nor does mention of trade names, commercial products, or organizations imply endorsement by the U.S. Government. This work was partially supported by the National Institutes of Health grant DK129539.

\bibliographystyle{apalike}
\bibliography{refbib}

\end{document}